\documentclass{article}

\usepackage[preprint]{neurips_2019}

\usepackage[utf8]{inputenc} % allow utf-8 input
\usepackage[T1]{fontenc}    % use 8-bit T1 fonts
\usepackage{xr}
\usepackage{url}            % simple URL typesetting
\usepackage{booktabs}       % professional-quality tables
\usepackage{amsfonts}       % blackboard math symbols
\usepackage{nicefrac}       % compact symbols for 1/2, etc.
\usepackage{microtype}      % microtypography
\usepackage{graphicx}
\usepackage{graphbox}
\usepackage{algorithm}
\usepackage{amsthm}
\usepackage[noend]{algpseudocode}
\usepackage{gensymb}
\usepackage{multirow}
\usepackage[hide=true,setmargin=true,marginparwidth=1.2in]{marginalia}

\usepackage{makecell}
\usepackage{amsbsy}

\newcommand{\comment}[1]{}

\newcommand{\Reals}{\mathbb{R}}

\title{\NA{Stabilizing the Lottery Ticket Hypothesis}\thanks{This article has been subsumed by \emph{Linear Mode Connectivity and the Lottery Ticket Hypothesis} \citep{frankle2019lottery}. Please read/cite that article instead.}}

\author{%
 Jonathan Frankle \\
 MIT CSAIL \\
 \And
 Karolina Dziugaite \\
 University of Cambridge \\
 Element AI \\
 \And
 Daniel M. Roy \\
 University of Toronto \\
 Vector Institute \\
 \And
 Michael Carbin \\
 MIT CSAIL \\
}

\begin{document}

\maketitle

\begin{abstract}
Pruning is a well-established technique for removing unnecessary structure from neural networks \NA{after training} to improve the performance of inference.
Several recent results have explored the possibility of pruning at initialization time to provide similar benefits during training.
In particular, the \emph{lottery ticket hypothesis} conjectures that typical neural networks contain small subnetworks that can train to similar accuracy in a commensurate number of steps.
The evidence for this claim is that a procedure based on iterative magnitude pruning (IMP) reliably finds such subnetworks retroactively on small vision tasks.
However, IMP fails on deeper networks, and proposed methods to prune before training or train pruned networks encounter similar scaling limitations.

In this paper, we argue that these efforts have struggled on deeper networks because they have focused on pruning precisely at initialization.
We modify IMP to search for subnetworks that could have been obtained by pruning \emph{early} in training (0.1\% to 7\% through) rather than at iteration 0.
With this change, it finds small subnetworks of deeper networks (e.g., 80\% sparsity on Resnet-50) that can complete the training process to match the accuracy of the original network on more challenging tasks (e.g., ImageNet).
In situations where IMP fails at iteration 0, the accuracy benefits of delaying pruning accrue rapidly over the earliest iterations of training.
To explain these behaviors, we study subnetwork \emph{stability}, finding that---as accuracy improves in this fashion---IMP subnetworks train to parameters closer to those of the full network and do so with improved consistency in the face of gradient noise.
These results offer new insights into the opportunity to prune large-scale networks early in training and the behaviors underlying the lottery ticket hypothesis.
\end{abstract}

\section{Introduction}

For decades, \emph{pruning} \citep{brain-damage, han-pruning} unnecessary structure from neural networks has been a popular way to improve the storage and computational costs of inference, which can often be reduced by an order of magnitude without harm to accuracy.
Pruning is typically a post-processing step after training;
until recently, it was believed that the pruned architectures could not themselves be trained from the start \citep{han-pruning, pruning-filters}.
New results challenge this wisdom, raising the prospect of reducing the cost of training by pruning beforehand.
\citet{rethinking-pruning} demonstrate that, at moderate levels of sparsity, pruning produces networks that can be reinitialized and trained to equal accuracy;
\citet{snip} propose an efficient method for finding such reinitializable subnetworks before training (SNIP).

\citet{lth} characterize the opportunity for pruning at initialization.
For shallow vision networks, they observe that---at levels of sparsity that are more extreme than \citet{rethinking-pruning} and \citet{snip}---pruned networks can successfully train from scratch so long as each unpruned connection is \emph{reset} back to its initial value from before training.\footnote{Appendix~\ref{app:comp} compares \citet{rethinking-pruning}, \citet{snip}, and \citet{lth}.}
This procedure, which we term \emph{iterative magnitude pruning}
(IMP; Algorithm \ref{fig:algorithm1} with $k=0$), produces a subnetwork of the original, untrained network; when it matches the accuracy of the original network, it is called a \emph{winning ticket}.
Based on these results, \citeauthor{lth} propose the \emph{lottery ticket hypothesis}: dense neural networks contain sparse subnetworks capable of training to commensurate accuracy at similar speed.

\begin{algorithm}
    \small
    \caption{Iterative Magnitude Pruning (IMP) with rewinding to iteration $k$.}
    \begin{algorithmic}[1]
    \State Randomly initialize a neural network $f(x; m \odot W_0)$ with initial trivial pruning mask $m = 1^{|W_0|}$.
    \State Train the network for $k$ iterations, producing network $f(x; m \odot W_k)$.
    \State Train the network for $T-k$ further iterations, producing network $f(x; m \odot W_t)$. \PROBLEM{DR: Is $t$ an input?}
    \State Prune the remaining entries with the lowest magnitudes from $W_T$. That is, let $m[i] = 0$ if $W_T[i]$ is pruned.
    \State If satisfied, the resulting network is $f(x; m \odot W_T)$.
    \State Otherwise, reset $W$ to $W_k$ and repeat steps 3-5 iteratively, gradually removing more of the network.
    \end{algorithmic}
    \label{fig:algorithm1}
\end{algorithm}

Despite the enormous potential of pruning before training, none of this work scales beyond small vision benchmarks.
\citeauthor{snip} provides results only for Tiny ImageNet, a restricted version of ImageNet with 200 classes.
\citeauthor{rethinking-pruning} examine Resnet-50 on ImageNet, but accuracy declines when only 30\% of parameters are pruned.
\citeauthor{rethinking-pruning}, \citet{gale}, and \citeauthor{lth} themselves show that IMP fails on deeper networks.
To find winning tickets on deeper networks for CIFAR10, \citeauthor{lth} make bespoke changes to each network's learning schedule.
In this paper, we argue that such efforts to train pruned networks or prune before training have struggled on deeper networks because they have focused on doing so precisely at initialization.

In comparison, other techniques gradually prune networks throughout training to competitive levels of sparsity without compromising accuracy \citep{zhu2017prune, gale, lym2019prunetrain, exploring, l0-reg, exploring}.
However, these approaches must maintain much of the network for a large portion of training or do not scale beyond toy benchmarks.

\textbf{Rewinding.}
\fTBD{JF: This begs the question: does SNIP work slightly later in training? We know it doesn't and we have evidence to back up that claim.}
In this paper, we demonstrate that there exist subnetworks of deeper networks (i.e., Resnet-50, Squeezenet, Inception-v3) at early points in training (0.1\% to 7\% through) that are 50\% to 99\% smaller and that can complete the training process to match the original network's accuracy.
We show this by modifying IMP to \emph{rewind} pruned subnetwork weights to their former values at iteration $k$ rather than \emph{resetting} them to iteration 0.
For networks where IMP cannot find a winning ticket, the accuracy benefits of this delay in pruning accrue rapidly over the earliest iterations of training. 
For example, IMP finds 80\% sparse subnetworks of Resnet-50 at epoch 6 (out of 90) with no loss in accuracy on ImageNet.
To the best of our knowledge, our work is the first to show that it is possible to prune (1) so early in training (2) to such extreme levels of sparsity (3) on such large-scale tasks.

\textbf{Stability.}
To explain why IMP fails when resetting to iteration 0 and improves rapidly when rewinding later, we introduce subnetwork \emph{stability}: the distance between two trained copies of the same subnetwork subjected to different noise.
In particular, we focus on the noise introduced by pruning (comparing the trained weights of the full network and subnetwork) and data order (comparing the weights of two subnetworks trained with different data orders).
In cases where IMP fails to find a winning ticket when resetting to iteration 0, both forms of stability improve rapidly as pruning is delayed during the early part of training, mirroring the rise in accuracy.
Stability to pruning captures the extent to which the subnetwork arrived at the same destination as the original network in the optimization landscape.
We hypothesize that improved stability to pruning means that a subnetwork comes closer to the original optimum and thereby accuracy; improvements in stability to data order mean the subnetwork can do so consistently in spite of the noise intrinsic to SGD.

Finally, we revise the lottery ticket hypothesis to consider rewinding in accordance with these results:

\newtheorem*{lth}{The Lottery Ticket Hypothesis with Rewinding}

\begin{lth}
  Consider a dense, randomly-initialized neural network $f(x; W_0)$ that trains to accuracy $a^*$ in $T^*$ iterations.
  Let $W_t$ be the weights at iteration $t$ of training.
  There exist an iteration $k \ll T^{*}$ and fixed pruning mask $m \in \{0, 1\}^{|W_0|}$ (where $||m||_1 \ll |W_0|$) such that subnetwork $m \odot W_k$ trains to accuracy $a \geq a^*$ in $T \leq T^* - k$ iterations.
\end{lth}

Based on this new understanding of the lottery ticket hypothesis provided by our rewinding and stability experiments, we conclude that there are unexploited opportunities to prune large-scale networks early in training while maintaining the accuracy of the eventual trained networks.

\section{Stability at Initialization}
\label{sec:stability-at-zero}

\newcommand{\Alg}[3]{\mathcal{A}^{#3}(#1,#2)}
\newcommand{\masks}{\mathcal{M}}
\newcommand{\loss}{\ell}
\newcommand{\Risk}[2]{L_{#1}(#2)}
\newcommand{\EmpRisk}[1]{\Risk{S}{#1}}

\begin{figure}
\centering
\vspace{-1em}
\includegraphics[width=0.45\textwidth]{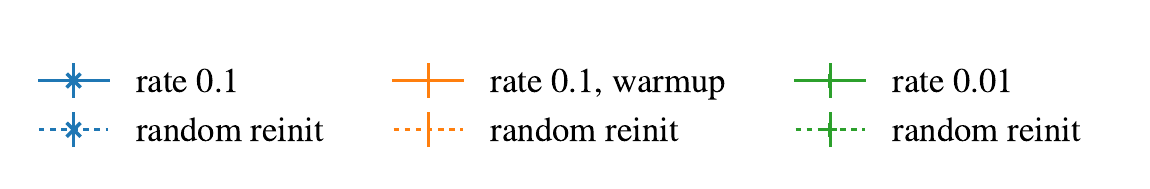}
\includegraphics[width=0.45\textwidth]{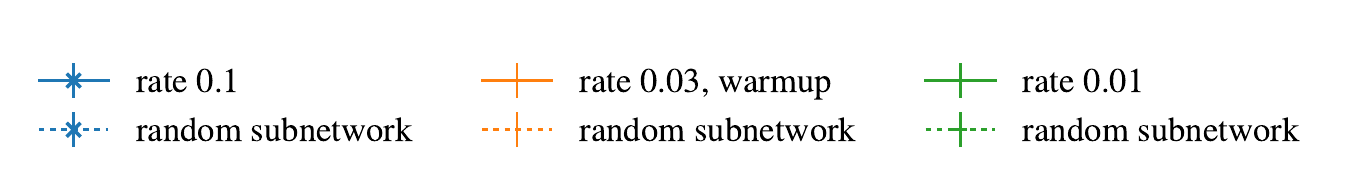}%
\vspace{-1em}
\includegraphics[width=0.45\textwidth]{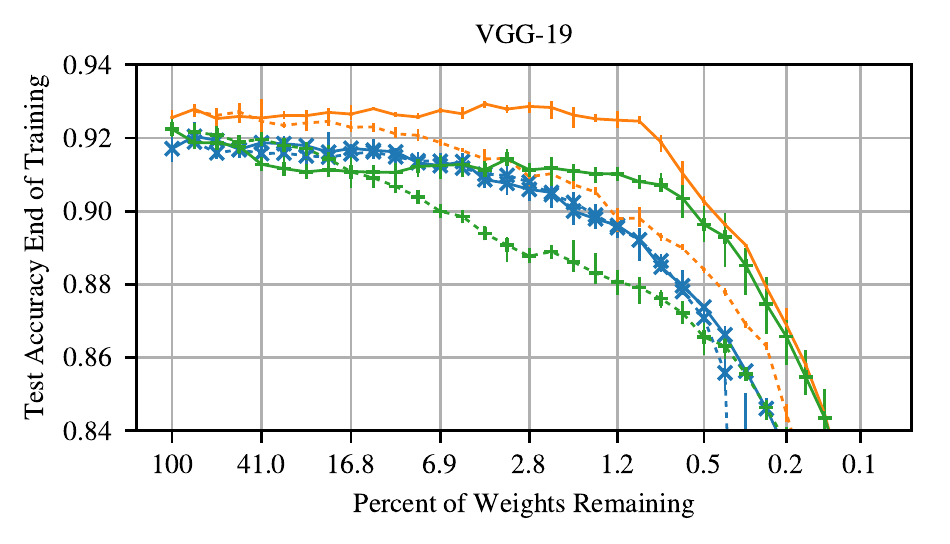}
\includegraphics[width=0.45\textwidth]{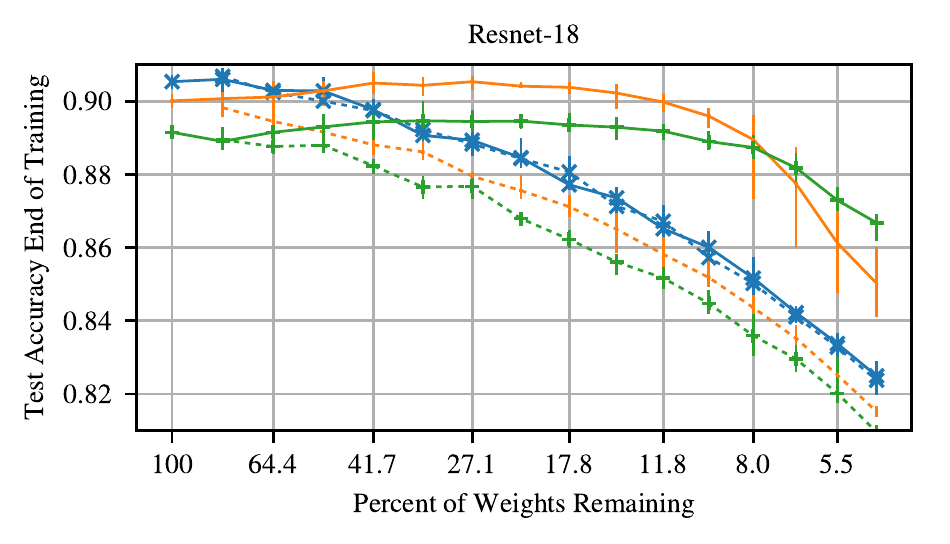}%
\vspace{-1em}
\caption{On deeper networks for CIFAR10, IMP fails to
 find winning tickets unless the learning rate schedule is altered.}
\label{fig:warmup}
\end{figure}

On deeper networks for CIFAR10, Iterative Magnitude Pruning (IMP; Algorithm~\ref{fig:algorithm1} at $k=0$) fails to yield winning tickets.
%Instead, the subnetworks that it finds perform no better than randomly reinitialized subnetworks.
The solid blue line in Figure \ref{fig:warmup} shows that IMP on VGG-19 (left) and Resnet-18 (right) produces no winning tickets;
the original initialization is inconsequential, and the networks can be reinitialized (dashed blue) without altering accuracy.
\citeauthor{lth} manage to find winning tickets in these networks by altering the learning schedule, lowering the learning rate (green) or adding warmup (orange).
However, they offer no principled justification for these choices, which are brittle and often alter the behavior of the unpruned networks.

To understand why IMP succeeds or fails to find winning tickets, we examine their \emph{stability}.\fTBD{MC: non-obvious transition}
We measure two forms of stability: 1) \emph{stability to pruning:} the distance between the weights of a subnetwork trained in isolation and the weights of the same subnetwork when trained within the larger network and 2) \emph{stability to data order:} the distance between the weights of two copies of a subnetwork trained with different data orders.
Stability to pruning captures a subnetwork's ability to train in isolation and still reach the same destination as the larger network. 
Stability to data order captures a subnetwork's intrinsic ability to consistently reach the same destination despite the gradient noise of SGD.
In this section, we demonstrate that, when IMP returns a subnetwork that qualifies as a winning ticket, it is dramatically more stable by both measures than a randomly-pruned network of the same size.

\textbf{Formal definitions.} A \emph{subnetwork} is a tuple $(W, m)$ of weights $W : \Reals^D$ and a fixed pruning mask $m : \{0, 1\}^D$. 
Notation $m \odot W$ denotes an element-wise product of a mask with weights.
A stochastic training \emph{algorithm} $\mathcal{A}^{t}: \Reals^D \times U \to \Reals^D$ maps initial weights $W$ and data order randomness $u \sim U$ to weights $W_t$ at iteration $t \in \{1,..,T\} $.
The distance $d$ between trained weights $W_{t}$ and $W'_{t}$ is the $L_2$ distance between the masked, trained parameters: $||m \odot W_t - m \odot W'_t||_2$. Throughout the paper, Appendix \ref{app:angle}, follows the same analysis for the angle between the masked, trained parameters.

The stability to pruning of a subnetwork $(W_t, m)$ with respect to a noise distribution $U$ under a distance $d(\cdot,\cdot)$ is the expected distance between masked weights at the end of training: $d(\Alg{W_t}{u}{T-t}, \Alg{m \odot W_t}{u}{T-t})$ for $u \sim U$.

The stability to data order of a subnetwork $(W_t, m)$ with respect to a noise distribution $U$ under a distance $d(\cdot,\cdot)$ is the expected distance between masked weights at the end of training: $d(\Alg{m \odot W_t}{ u}{T-t}, \Alg{m \odot W_t}{u'}{T-t})$ for $u, u' \sim U$. 

\begin{figure}
\scriptsize
\centering
\begin{tabular}{l | c@{\ \ }c@{\ \ } c@{\ \ }c@{\ \ }c@{\ \ }c@{\ \ }c@{\ \ } c@{\ \ } c@{\ \ }}
\toprule
Network & Dataset & Params& Iters & Batch & Accuracy & Rate & Schedule & Warmup & Winning? \\ \midrule
Lenet & MNIST & 266K & 50K & 60 & 98.0 $\pm$ 0.2\% & adam 12e-4 & constant & 0 & Y \\ \midrule
Resnet-18 (standard) &  &  &  &  & 90.5 $\pm$ 0.0\% & mom. 0.1 & \multirow{3}{*}{\makecell{10x drop at\\56K, 84K}}&  0 & N \\
Resnet-18 (low) & CIFAR10 & 274K & 30K & 128 & 89.1 $\pm$ 0.1\% & mom. 0.01 & & 0 & Y \\
Resnet-18 (warmup) &  & & & & 89.6 $\pm$ 0.1\% & mom. 0.03 & & 20K & Y \\ \midrule
VGG-19 (standard) & & & & & 91.5 $\pm$ 0.2\% & mom. 0.1 & \multirow{3}{*}{\makecell{10x drop at\\20K, 25K}} & 0 & N \\
VGG-19 (low) & CIFAR10 & 20.0M & 112K & 64& 92.0 $\pm$ 0.1\% & mom. 0.01 & &  0 & N \\
VGG-19 (warmup) & & & & & 92.3 $\pm$ 0.1\% & mom. 0.1 & & 10K & Y \\
\bottomrule
\end{tabular}
\caption{Networks for MNIST and CIFAR10. Accuracies averaged across three trials.}
\label{fig:small-networks}
\end{figure}

\begin{figure}
\centering
\scriptsize
\begin{tabular}{l@{\ \ }| @{\ \ }c@{\ \ } | c@{\ \ \ \ }c@{\ \ \ \ }c | c@{\ \ \ \ }c@{\ \ \ \ }c | c@{\ \ \ \ }c }
\toprule
Network & Sparsity & \multicolumn{3}{c|}{Data Order Stability (Distance)} & \multicolumn{3}{c|}{Pruning Stability (Distance)} & \multicolumn{2}{c}{Accuracy} \\
& & IMP & Random & Comp & IMP & Random & Comp & IMP & Random \\  \midrule
Lenet  & 10.7\% & 20.7 $\pm$ 0.6 & 58.6 $\pm$ 4.0 & 2.8x & 48.1 $\pm$ 0.6 & 75.7 $\pm$ 0.5 & 1.6x & 98.3 $\pm$ 0.1 & 97.5 $\pm$ 0.3 \\ \midrule
Resnet-18 (standard) & & 66.4 $\pm$ 0.7 & 66.7 $\pm$ 1.1 & 1.0x & 54.4 $\pm$ 0.2 & 53.4 $\pm$ 0.4 & 1.0x & 87.7 $\pm$ 0.4 & 87.7 $\pm$ 0.5 \\
Resnet-18 (low) & 16.7\% &  \hphantom{0}7.1 $\pm$ 1.2& 28.9 $\pm$ 3.2 & 4.1x & 19.8 $\pm$ 1.4 & 26.6 $\pm$ 0.1 & 1.3x &  89.1 $\pm$ 0.4 & 86.1 $\pm$ 0.6 \\
Resnet-18 (warmup) &  & \hphantom{0}9.5 $\pm$ 0.1 & 37.4 $\pm$ 1.9 & 3.9x & 24.8 $\pm$ 0.9 & 34.6 $\pm$ 0.2 & 1.4x & 90.3 $\pm$ 0.4 & 86.8 $\pm$ 0.5 \\ \midrule
VGG-19 (standard) & & 285 $\pm$ 3 & 245 $\pm$ 34 & 0.8x & 216 $\pm$ 1 & 212 $\pm$ 1 & 1.0x & 90.0 $\pm$ 0.3 & 90.2 $\pm$ 0.5 \\
VGG-19 (low) & 2.2\% & 36.8 $\pm$ 2.6 & 90.3 $\pm$ 0.7 & 2.5x & 44.0 $\pm$ 0.3 & 66.1 $\pm$ 0.4 & 1.5x & 91.0 $\pm$ 0.3 & 88.0 $\pm$ 0.3 \\
VGG-19 (warmup) & & 97 $\pm$ 0.6 & 267 $\pm$ 2 & 2.7x &  138 $\pm$ 1 & 201 $\pm$ 1 & 1.5x & 92.4 $\pm$ 0.2 & 90.3 $\pm$ 0.3 \\
\bottomrule
\end{tabular}
\caption{The average data order stability of subnetworks obtained by IMP and by randomly pruning. Errors are the minimum or maximum across 18 samples.}
\label{fig:stability-at-zero}
\end{figure}

\textbf{Methodology.}
We measure both forms of stability for Lenet for MNIST and Resnet-18 and VGG-19 for CIFAR10 as studied by \citeauthor{lth} and described in Figure \ref{fig:small-networks}. We do so for networks produced by IMP as well as randomly-pruned networks in which we generate a random mask $m$ of a given size with the same layer-wise proportions.
These networks are trained for a fixed number of epochs.
During each epoch, all training examples are randomly shuffled, randomly augmented, and separated into minibatches without replacement; network parameters are updated based on each minibatch in sequence until the training data is exhausted, after which the next epoch begins. 

\textbf{Results.}
Figure \ref{fig:stability-at-zero} displays the stability for IMP subnetworks at a representative level of sparsity.
IMP finds winning tickets for Lenet, but cannot do so for Resnet-18 or VGG-19 without warmup.
Whenever it does so, the winning ticket is far more stable than a randomly-sampled subnetwork.
For example, Lenet winning tickets are 2.8x (data order) and 1.6x (pruning) closer in $L_2$ distance than random subnetworks.
Conversely, IMP cannot find a winning ticket within VGG-19 (standard) and Resnet-18 (standard), and the subnetworks it returns are no more stable than random subnetworks.%

\textbf{Discussion.}
These results suggest a connection between winning tickets and stability. Winning tickets are far more stable than random subnetworks.
Likewise, when IMP subnetworks are no more accurate than random subnetworks, they are correspondingly no more stable.
Interestingly, \citeauthor{lth}'s techniques that enable IMP to find winning tickets also benefit stability.

\section{Stability with Rewinding}
\label{sec:rewinding}

\begin{figure}
\vspace{-.4em}%
\includegraphics[width=0.25\textwidth]{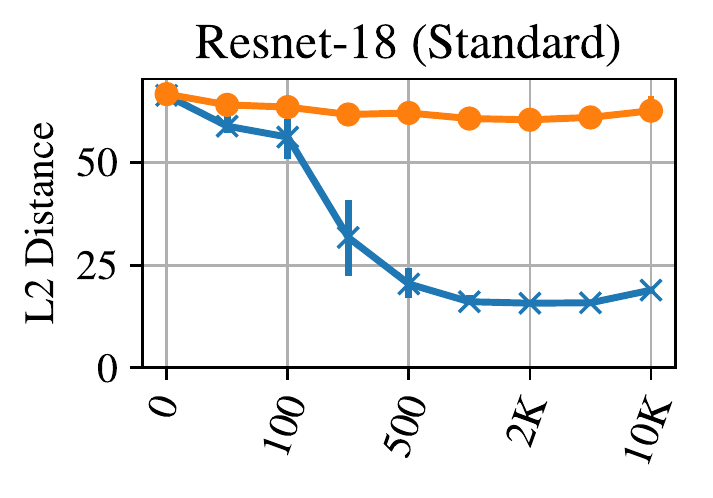}%
\includegraphics[width=0.25\textwidth]{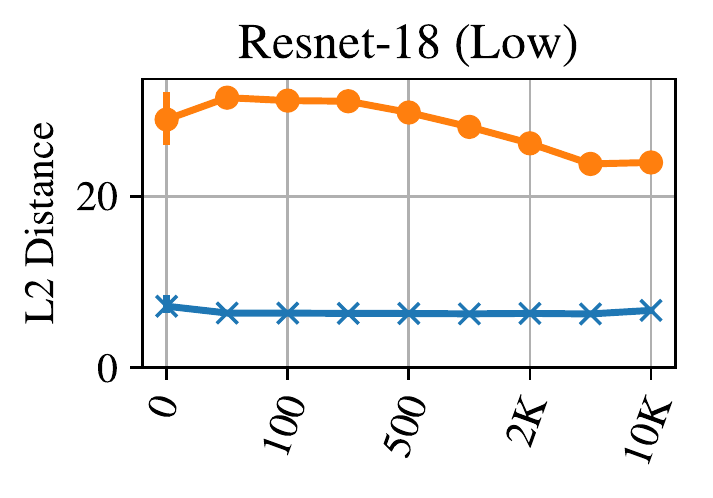}%
\includegraphics[width=0.25\textwidth]{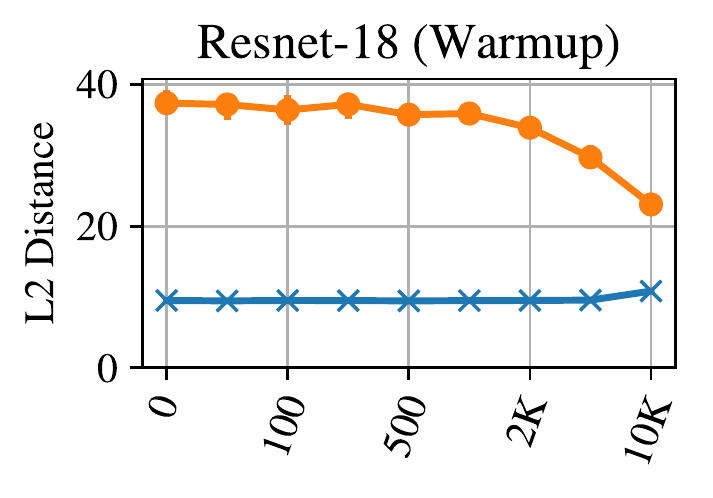}%
\includegraphics[width=0.25\textwidth]{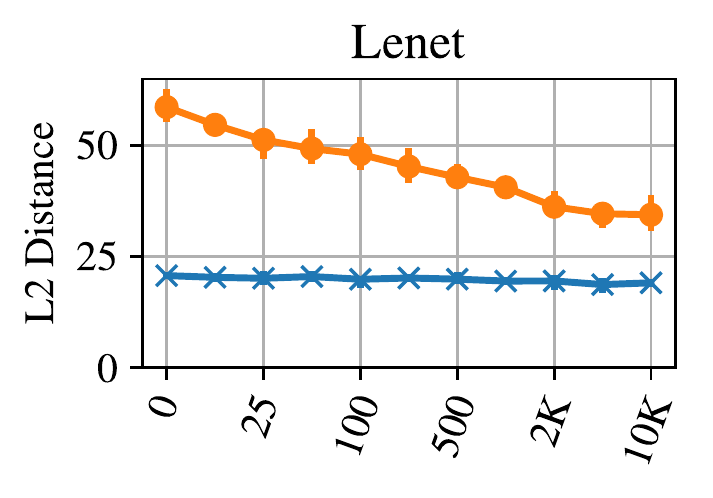}

\vspace{-.4em}%
\includegraphics[width=0.25\textwidth]{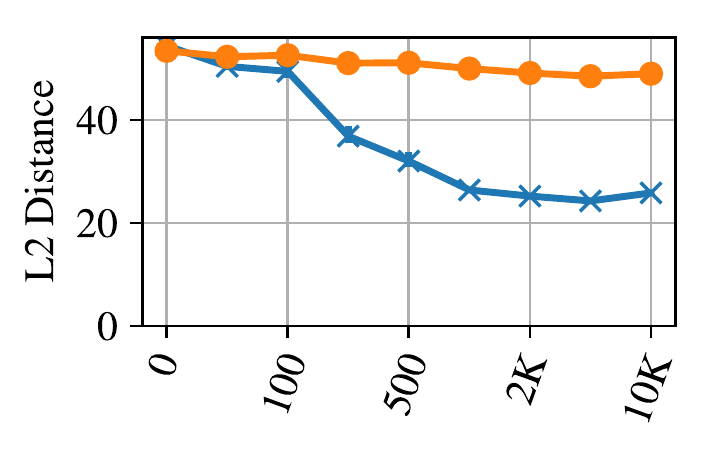}%
\includegraphics[width=0.25\textwidth]{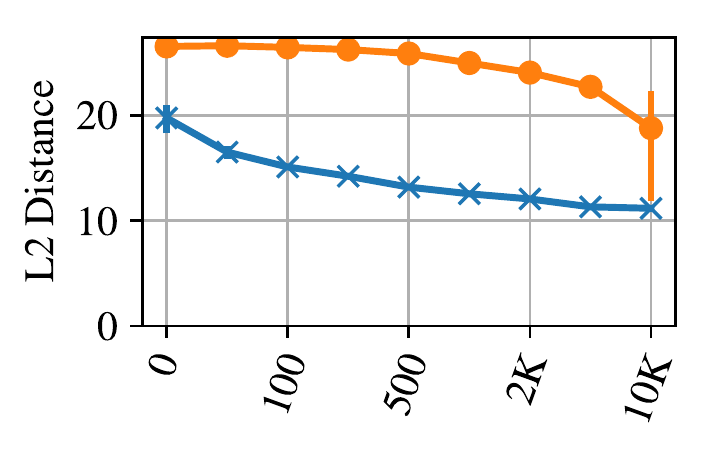}%
\includegraphics[width=0.25\textwidth]{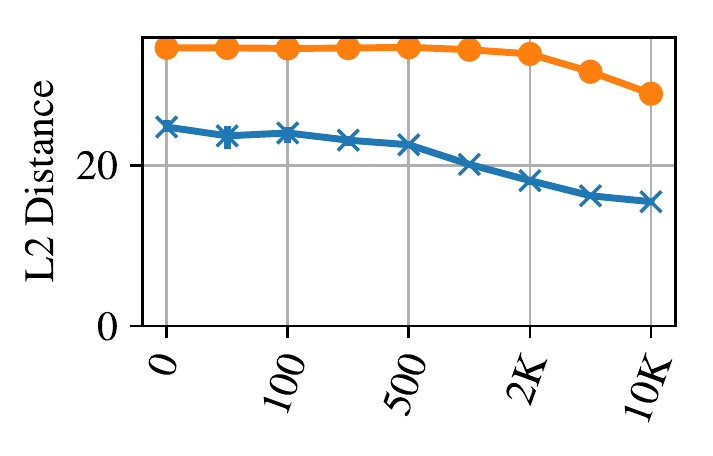}%
\includegraphics[width=0.25\textwidth]{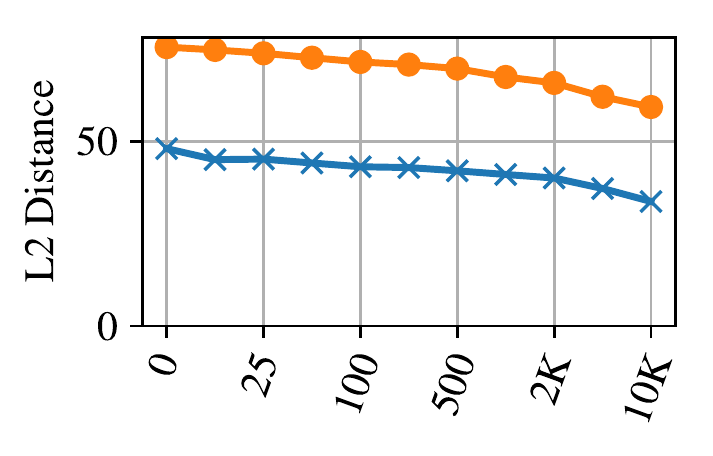}

\vspace{-.6em}%
\includegraphics[width=0.25\textwidth]{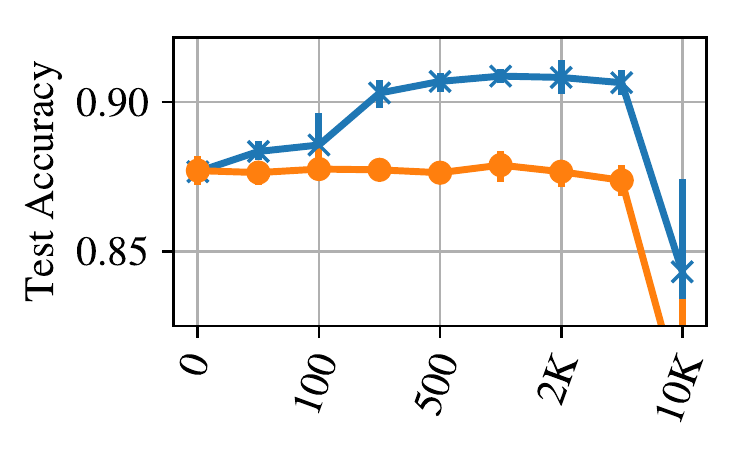}%
\includegraphics[width=0.25\textwidth]{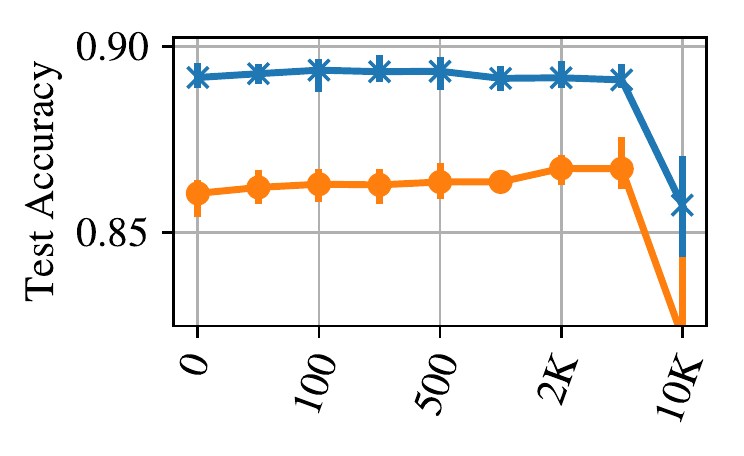}%
\includegraphics[width=0.25\textwidth]{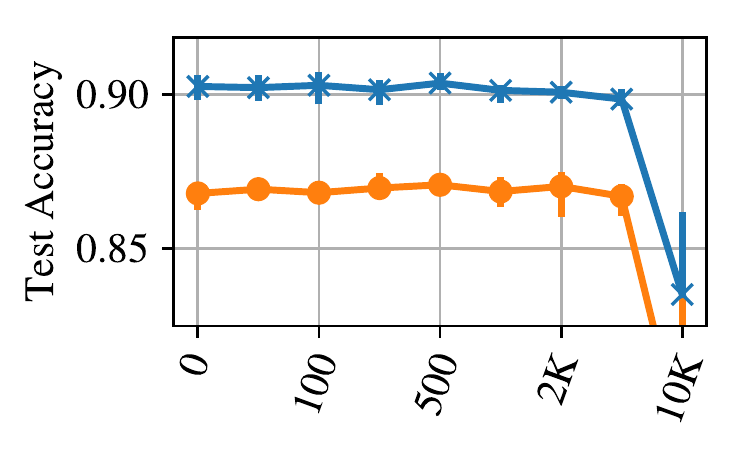}%
\includegraphics[width=0.25\textwidth]{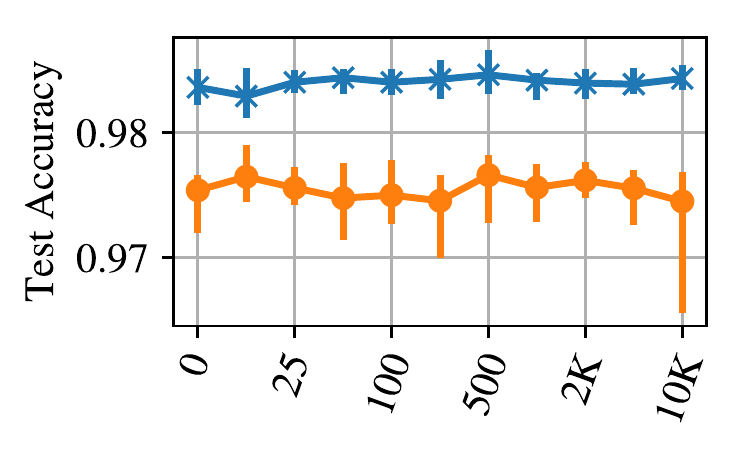}

\vspace{-.6em}%
\includegraphics[width=0.25\textwidth]{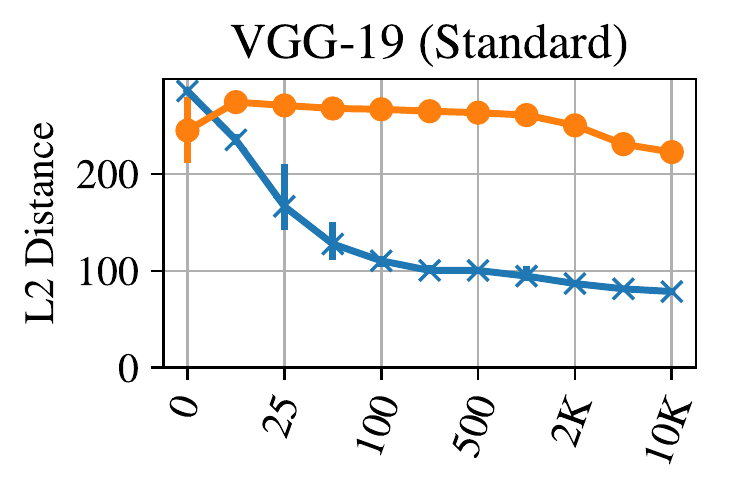}%
\includegraphics[width=0.25\textwidth]{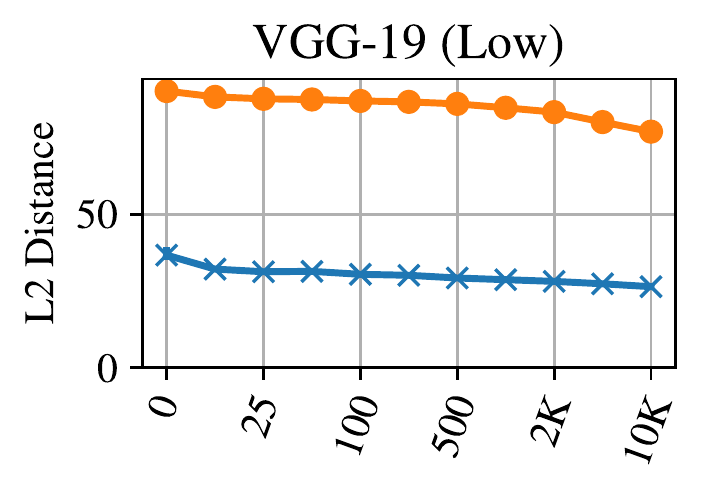}%
\includegraphics[width=0.25\textwidth]{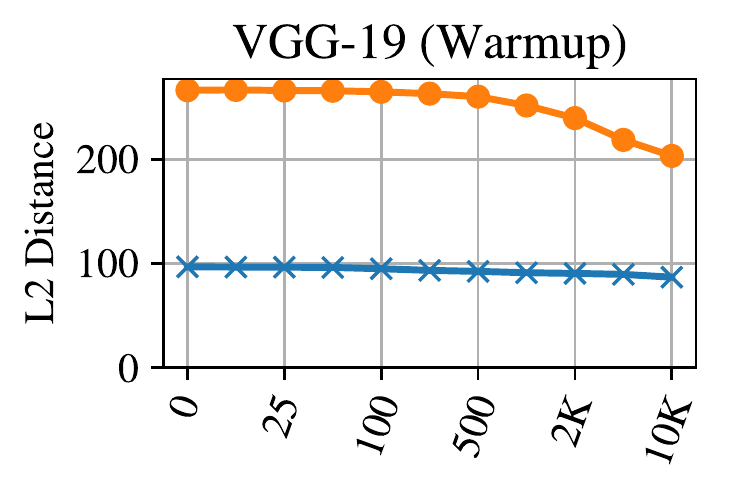}%
\hspace{.05\textwidth}\includegraphics[width=.2\textwidth]{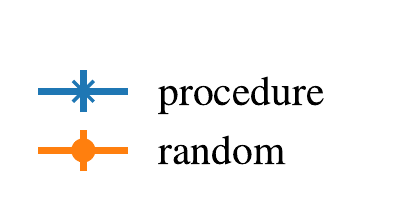}

\vspace{-.6em}%
\includegraphics[width=0.25\textwidth]{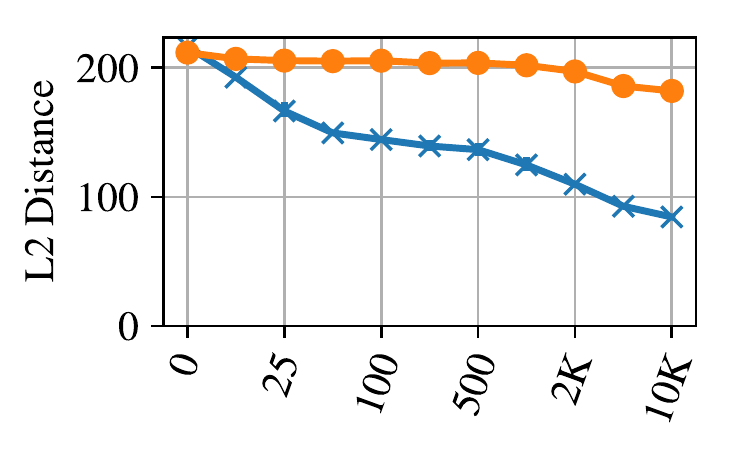}%
\includegraphics[width=0.25\textwidth]{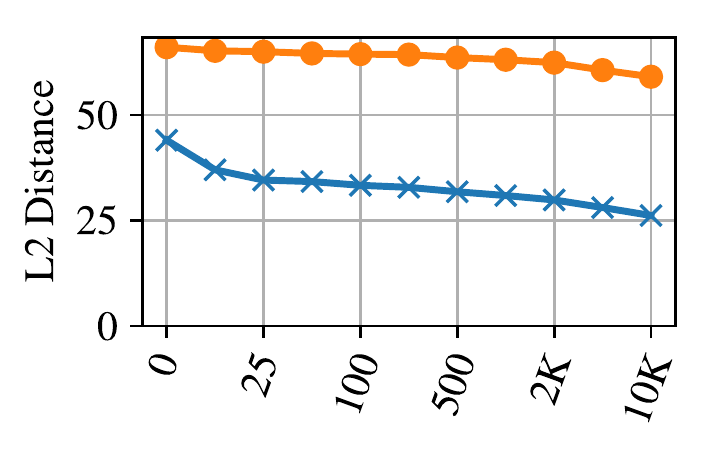}%
\includegraphics[width=0.25\textwidth]{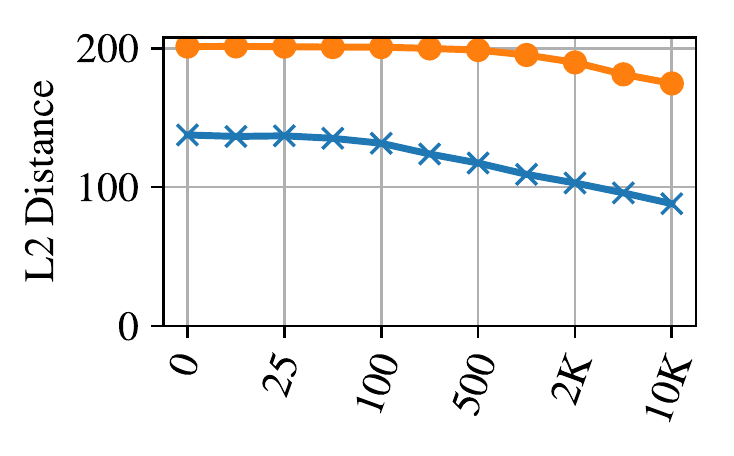}%

\vspace{-.6em}%
\includegraphics[width=0.25\textwidth]{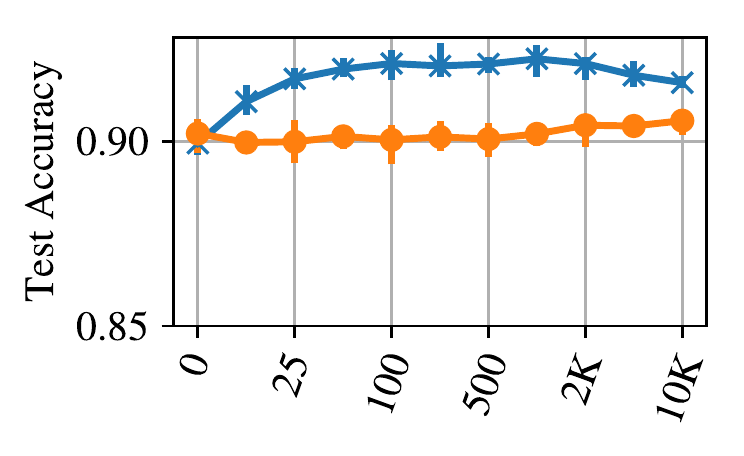}%
\includegraphics[width=0.25\textwidth]{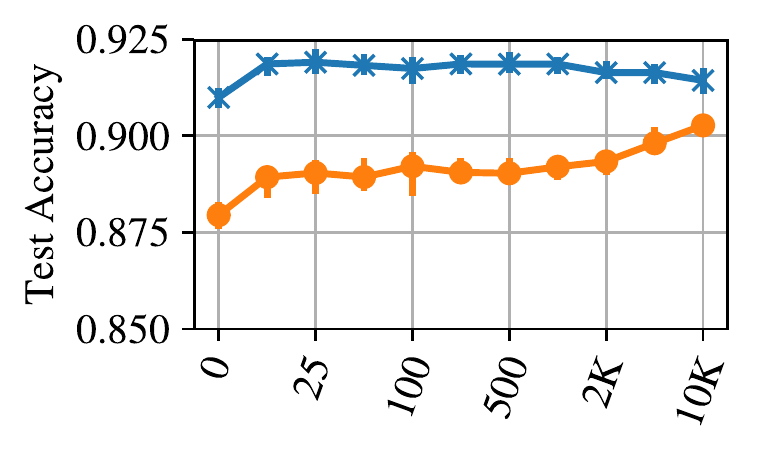}%
\includegraphics[width=0.25\textwidth]{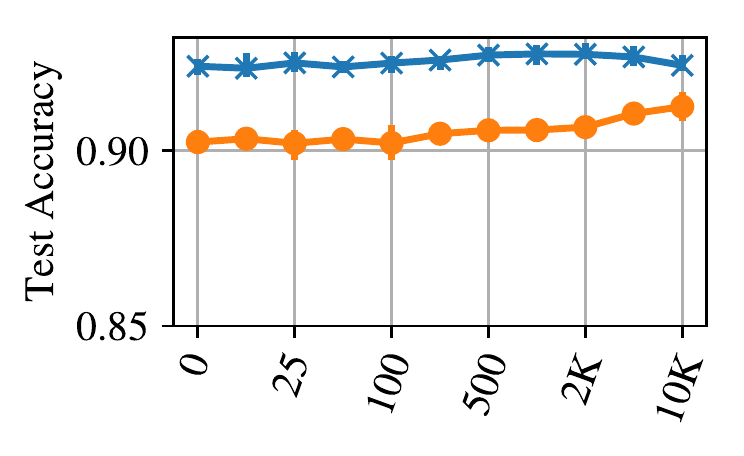}%
\caption{The effect of the rewinding iteration (x-axis) on data order (top row) and pruning (middle row) stability and accuracy (bottom row) for networks in Figure \ref{fig:small-networks}. Error bars are the minimum and maximum across 18 (data order) and three samples (stability and accuracy). These results were obtained by training more than 6500 networks on K80s and V100s.}
\label{fig:rewinding}
\end{figure}

In cases where the IMP fails to find a winning ticket, reducing the learning rate early in training via warmup makes it possible for the procedure to succeed, simultaneously improving the stability of the resulting subnetwork.
The efficacy of warmup suggests that a high learning rate is not necessarily detrimental later in training---only in the initial iterations.
One hypothesis to explain these results is that---for the less overparameterized regime of training a subnetwork---optimization is particularly sensitive to noise in the early phase of training.
This sensitivity can be mitigated by reducing the learning rate early on, minimizing the consequences of any individual misstep.

Under this hypothesis, we would expect these subnetworks to become more stable later in training, when they better tolerate a high learning rate.
We explore this possibility by modifying IMP (Algorithm \ref{fig:algorithm1} with $k \neq 0$).
After finding a subnetwork, \emph{rewind} each connection back to its weight from an earlier iteration $k$ rather than \emph{resetting} it back to its initial value as \citeauthor{lth} do.

\textbf{Results.}
The upper two plots for each network in Figure \ref{fig:rewinding}\fTBD{MC: same max value of y-axis for all plots for a given network.} show the stability (measured in distance) of the IMP subnetwork (blue) and a random subnetwork (orange) across rewinding iterations.
Across all networks, rewinding later causes gradual improvements in stability, supporting our hypothesis.

For Resnet-18 (standard) and VGG-19 (standard), in which no winning tickets can be found, IMP subnetworks are no more stable than randomly-pruned subnetworks when reset to iteration 0.
However, as the rewinding iteration increases, the IMP subnetwork's stability dramatically improves when compared to its stability at iteration 0.
Up to our point of analysis, this improvement is larger than that for random subnetworks.
For IMP subnetworks, this change takes place rapidly---within the first 100 iterations (0.14 epochs) for VGG-19 and 500 iterations (1.4 epochs) for Resnet-18.

In cases where IMP finds a winning ticket, IMP subnetworks are already more stable than random subnetworks when resetting to iteration 0 (as noted in Section \ref{sec:stability-at-zero}).
This stability gap remains in place across all rewinding iterations that we consider, although it gradually shrinks toward the end of our range of analysis as the random subnetworks become somewhat more stable;
by this point, the networks have already made substantial training progress or---in the case of Lenet---converged.

\textbf{Discussion.}
Section \ref{sec:stability-at-zero} shows that, when IMP identifies a winning ticket, it is more stable than a random subnetwork.
Here, we find that IMP subnetworks generally achieve such a stability gap across all network configurations;
however, in many cases, this occurs a small way into training.

\section{Accuracy with Rewinding}
\label{sec:rewinding-acc}

Section \ref{sec:stability-at-zero} observes that winning tickets found by IMP exhibit both improved accuracy and stability over random subnetworks.
The previous section finds that the stability of IMP subnetworks improves as a function of rewinding iteration, especially rapidly so for networks where winning tickets are not found.
The bottom plots in Figure \ref{fig:rewinding} show that accuracy improves in a similar manner.
Although resetting to iteration 0 does not produce winning tickets for Resnet-18 and VGG-19, rewinding slightly later (iteration 500 for Resnet-18 and 100 for VGG-19) leads to IMP subnetworks that exceed the accuracy of the original network.
At later rewinding iterations, the rate of stability improvements subsides for these subnetworks, corresponding to slowed or no improvement in accuracy.

Improved stability also results in improved accuracy for random subnetworks.
However, the accuracy of the random subnetworks remains lower in all of our experiments.
We believe that the significant stability gap explains this difference:
we hypothesize that the stability of IMP subnetworks over their random counterparts results in better accuracy.

For the latest rewinding iterations that we consider, IMP subnetwork accuracy declines in many cases.
According to our methodology, we train a subnetwork rewound to iteration $k$ for $T^* - k$ iterations, where $T^*$ is the iterations for which the original network was trained.
At later rewind points, we believe that this does not permit enough training time for the subnetwork to recover from pruning.

\begin{figure}
\centering
\vspace{-1em}
\includegraphics[width=0.45\textwidth]{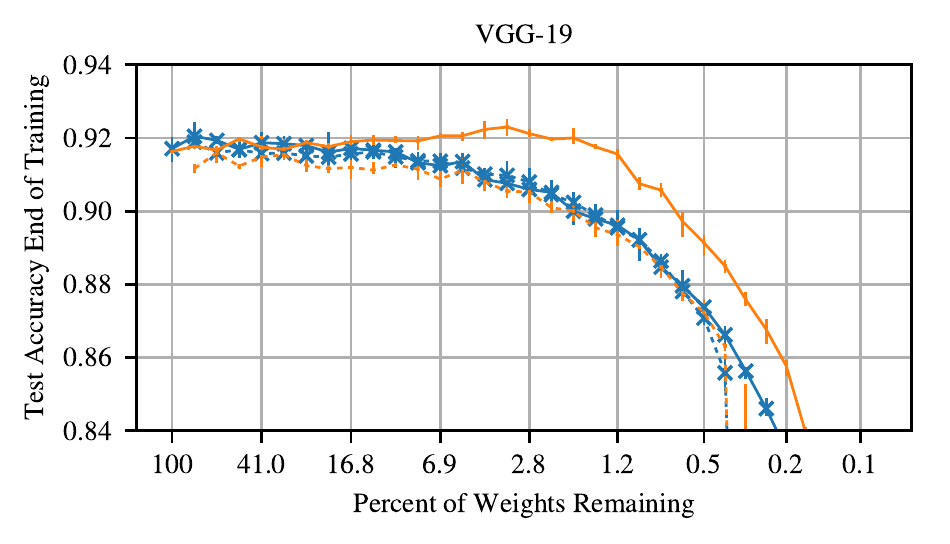}%
\includegraphics[width=0.45\textwidth]{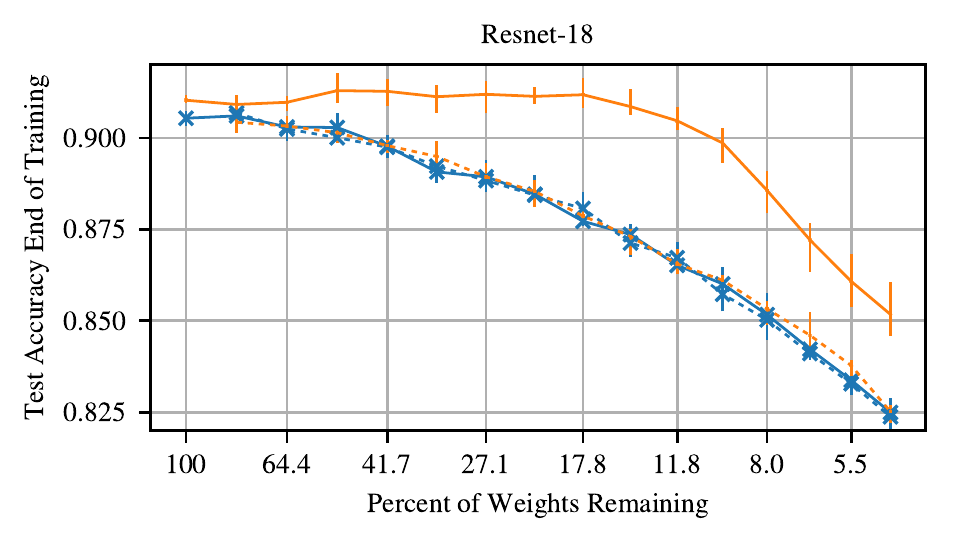}%
\vspace{-1em}
\includegraphics[width=0.35\textwidth]{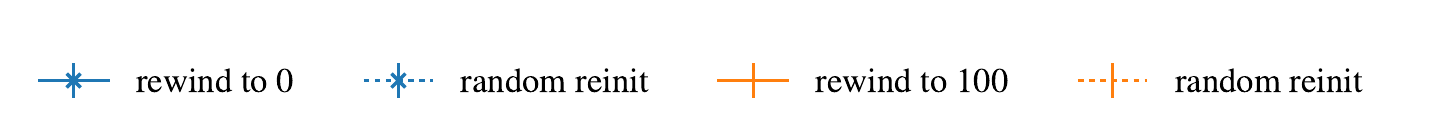}%
\includegraphics[width=0.35\textwidth]{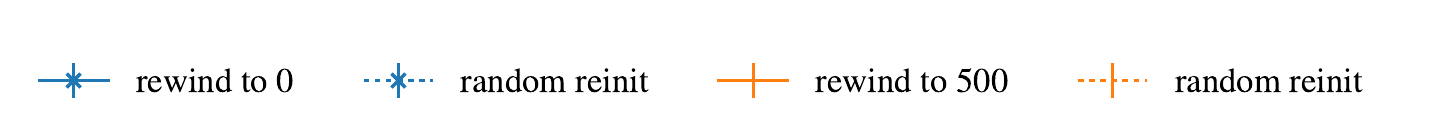}%
\vspace{-1em}
\caption{IMP subnetworks rewound to an iteration early in training outperform the original networks, while resetting to iteration 0 does not.}
\label{fig:small-late}
\end{figure}

\textbf{Discussion.}
Figure \ref{fig:small-late} plots the accuracy of Resnet-18 and VGG-19 across all levels of sparsity, comparing the IMP subnetworks reset to iteration 0 with those rewound to iteration 100 (VGG-19) and iteration 500 (Resnet-18)---the iterations at which stability and accuracy improvements saturate.
At all levels of sparsity, rewinding makes it possible to find trainable subnetworks early in training without any modifications to network hyperparameters (unlike the warmup or reduced learning rates required in Figure \ref{fig:warmup}).

These findings reveal a new opportunity to improve the performance of training.
The aspiration behind the lottery ticket hypothesis is to find small, trainable subnetworks before any training has occurred.
Insofar as IMP reflects the extent of our knowledge about the existence of equally-capable subnetworks early in training,
our findings suggest that---for deeper networks---the best opportunity to prune is a small number of iterations into training rather than at initialization.
Doing so would exploit the rapid improvements in subnetwork stability and accuracy, resulting in subnetworks that can match the performance of the original network at far greater levels of sparsity.

\section{Rewinding on Deep Networks for ImageNet}
\label{sec:imagenet}

\begin{figure}
\centering
\scriptsize
\begin{tabular}{l | c@{\ \ }c@{\ \ }c@{\ \ }c@{\ \ }l@{\ \ }c}
\toprule
Network & Params& Eps. & Batch & Top-1 Accuracy & Learning Schedule & Sparsity\\ \midrule
Resnet-50 & 25.5M & 90 & 1024 & 76.1 $\pm$ 0.1\% & rate 0.4; warmup 5 eps.; 10x drop at 30, 60, 80; momentum & 70\% \\
Inception-v3 & 27.1M & 171 & 1024 & 78.1 $\pm$ 0.1\% & rate 0.03 linearly decayed to 0.005; momentum & 70\%\\
Squeezenet & 1.25M & 150 & 1024 & 54.8 $\pm$ 0.6 \% & rate 0.66 exponentially decayed to 6.6e-5; rmsprop & 50\% \\
\bottomrule
\end{tabular}
\caption{Networks for experiments on ImageNet with TPUs. Accuracies averaged across five trials.}
\label{fig:large-networks}
\end{figure}

Rewinding made it possible to find sparse, trainable subnetworks of deeper networks for CIFAR10 without the need to alter the underlying network's hyperparameters.
In this section, we demonstrate that this strategy serves the same purpose for deeper networks for ImageNet \citep{imagenet}.
Although IMP with $k=0$ does not produce winning tickets, rewinding 4.4\%, 3.5\%, and 6.6\% into training yields subnetworks that are 70\%, 70\%, and 50\% smaller than the Resnet-50 \citep{resnet}, Inception-v3 \citep{inception2}, and Squeezenet \citep{squeezenet}  architectures, respectively, that can complete training without any drop in accuracy.
We trained more than 600 networks using standard implementations for TPUs \citep{tpu-implementation} as described in Figure \ref{fig:large-networks}.

\begin{figure}
\vspace{-.4em}%
\includegraphics[width=0.24\textwidth]{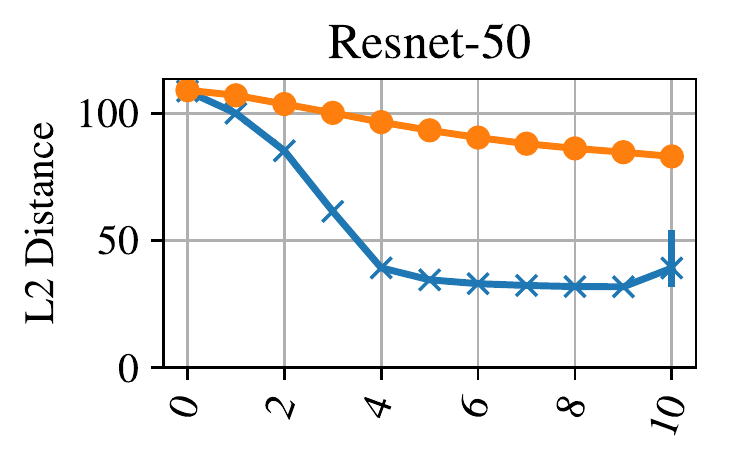}%
\includegraphics[width=0.24\textwidth]{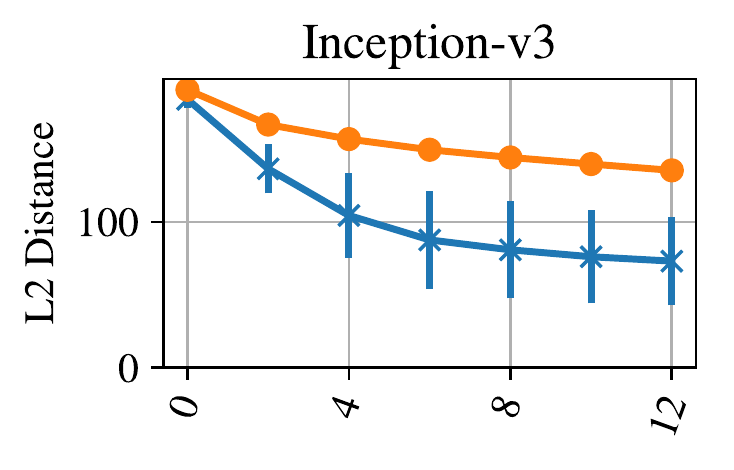}%
\includegraphics[width=0.24\textwidth]{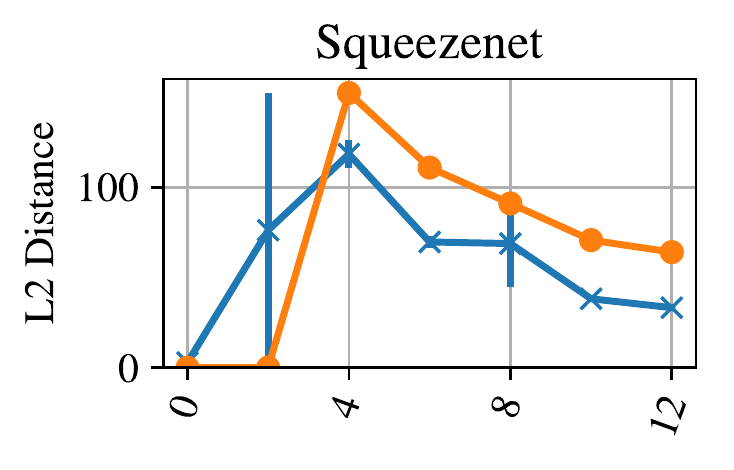}
\includegraphics[width=.2\textwidth]{figures/late/lenet-accuracy-dataorder-legend}\hspace{.03\textwidth}%

\vspace{-.6em}
\includegraphics[width=0.24\textwidth]{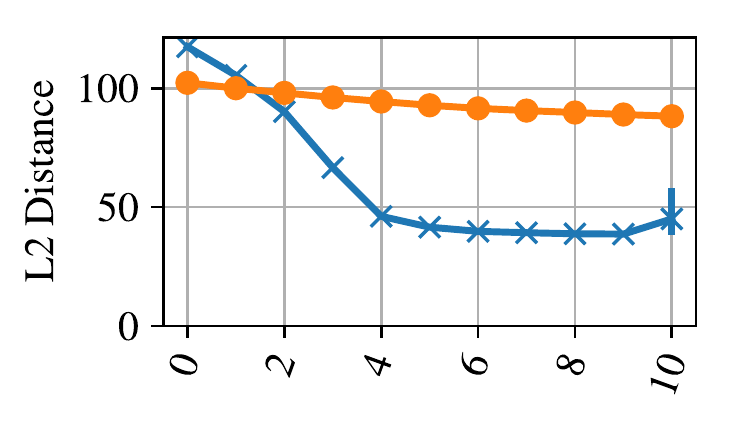}%
\includegraphics[width=0.24\textwidth]{figures/late/inception-parameter-dataorder}%
\includegraphics[width=0.24\textwidth]{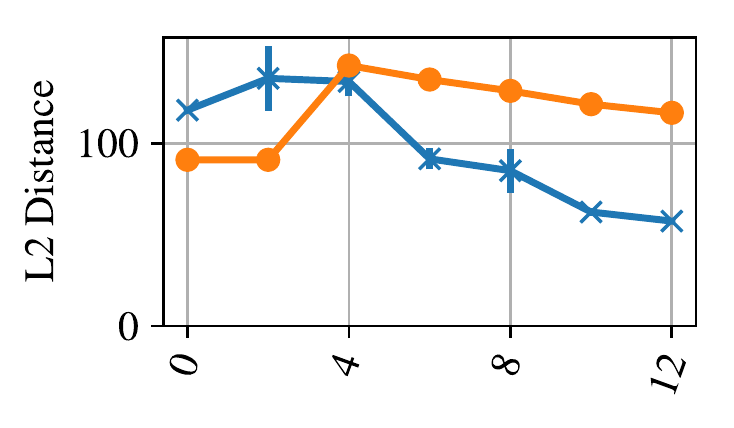}

\vspace{-.6em}%
\includegraphics[width=0.24\textwidth]{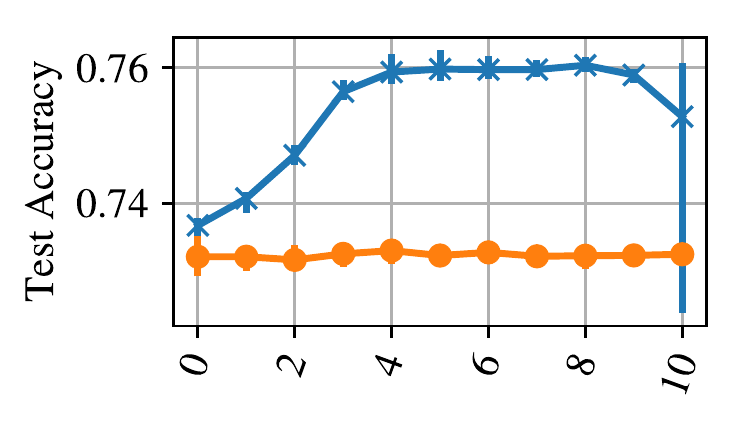}%
\includegraphics[width=0.24\textwidth]{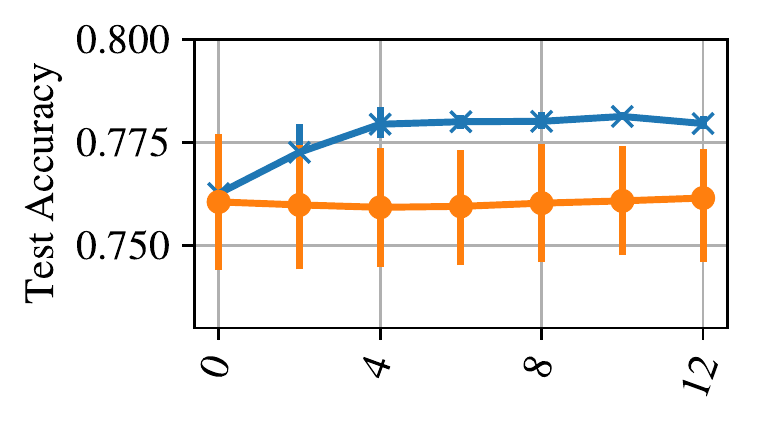}
\includegraphics[width=0.24\textwidth]{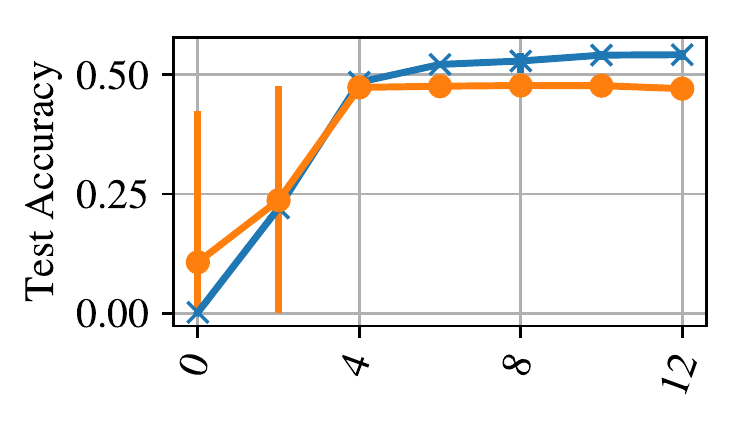}

\caption{The effect of the rewinding epoch (x-axis) on data order stability and accuracy.}
\label{fig:imagenet-late-resetting}
\end{figure}

Figure \ref{fig:imagenet-late-resetting} shows the effect of the rewinding iteration on the stability and accuracy at the levels of sparsity from Figure \ref{fig:large-networks}.
In general, the trends from Section \ref{sec:rewinding-acc} remain in effect.
When resetting weights to initialization, IMP subnetworks perform no better than random subnetworks in terms of both stability and accuracy.
As the rewinding iteration increases, a gap in stability emerges.
Accuracy improves alongside stability, reaching the accuracy of the original network at epoch 4 (out of 90) for Resnet-50, 6 (out of 171) for Inception-v3, and 10 (out of 150) for Squeezenet.
In the case of Squeezenet, rewinding too early makes it impossible for the subnetworks to learn at all.

\begin{figure}
\centering
\vspace{-1em}
\includegraphics[width=0.45\textwidth]{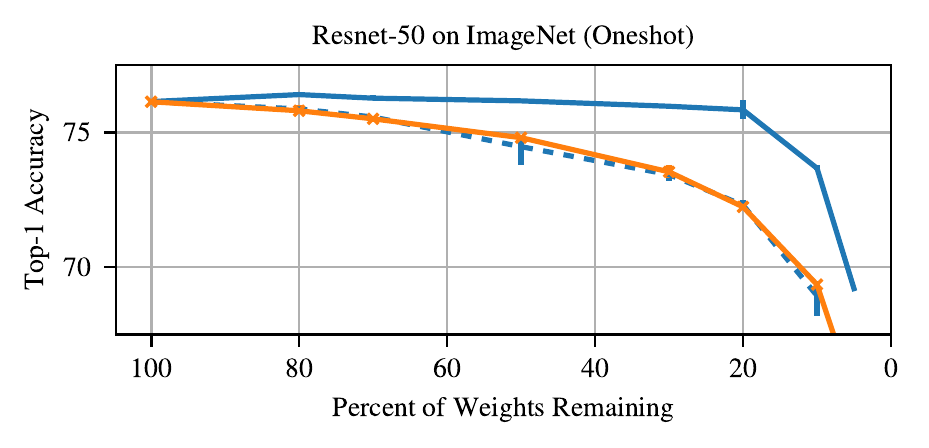}%
\includegraphics[width=0.45\textwidth]{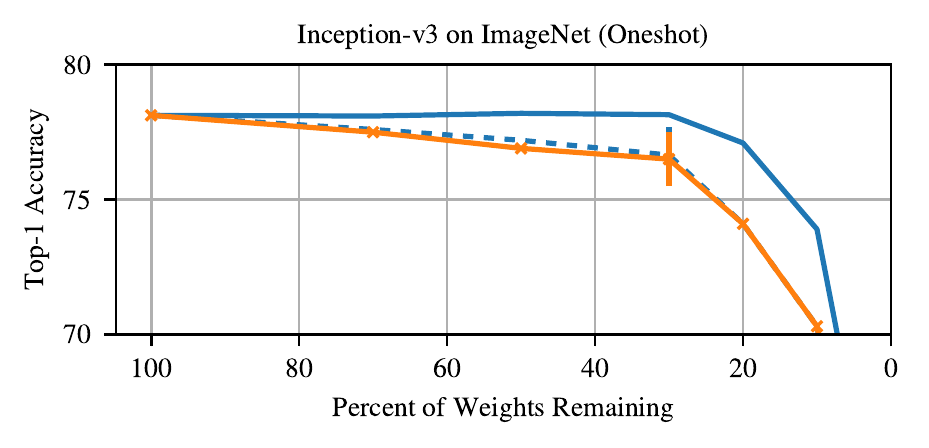}

\vspace{-1em}
\includegraphics[width=0.45\textwidth]{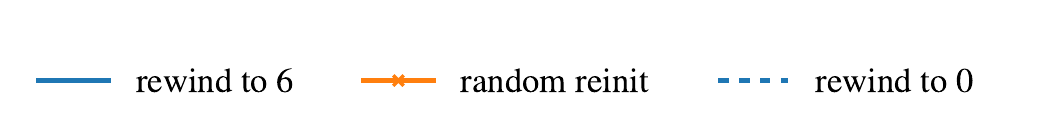}%
\includegraphics[width=0.45\textwidth]{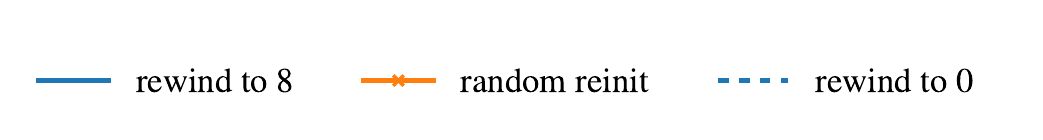}

\vspace{-1em}
\includegraphics[width=0.45\textwidth]{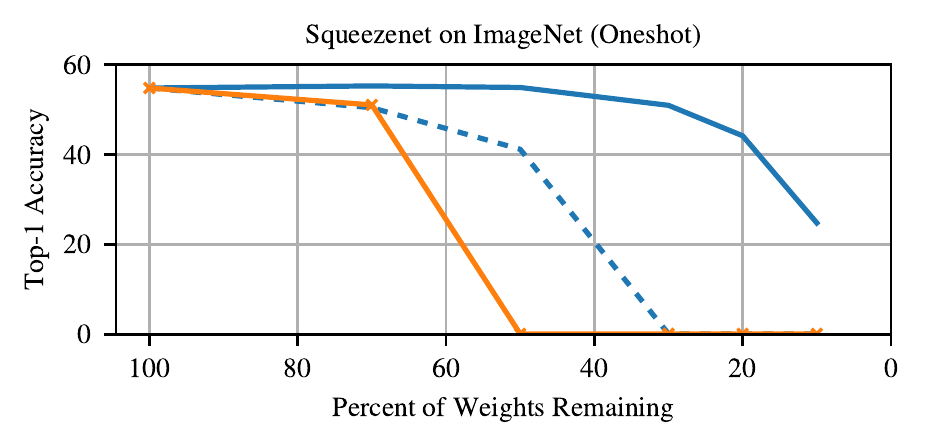}%
\includegraphics[width=0.45\textwidth]{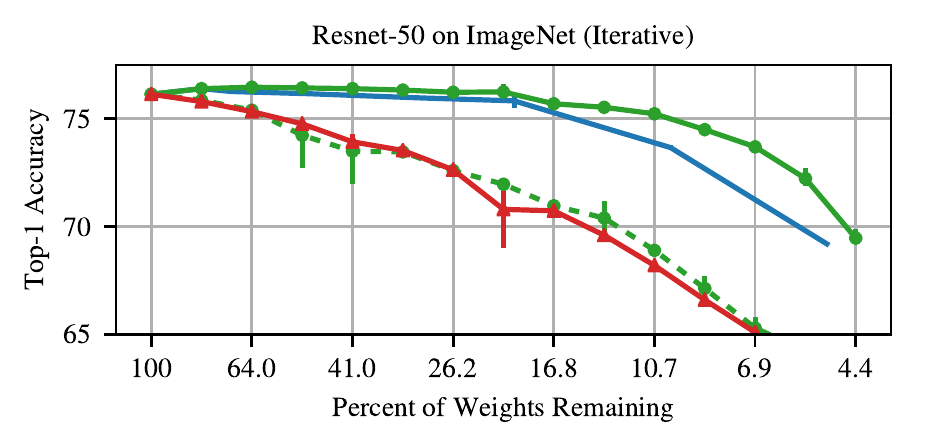}

\vspace{-1em}
\includegraphics[width=0.45\textwidth]{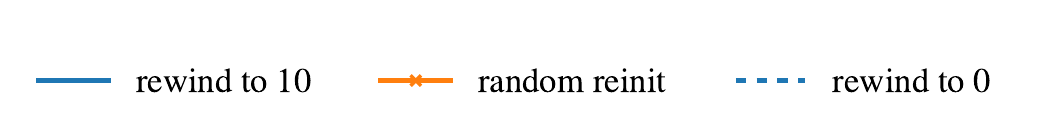}%
\includegraphics[width=0.45\textwidth]{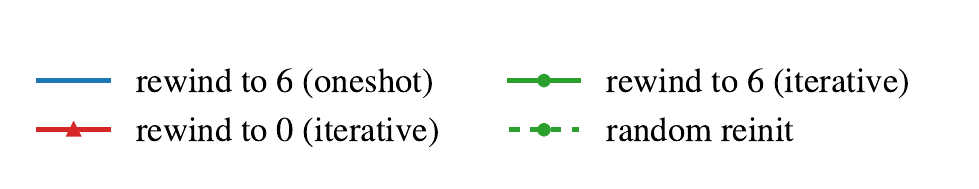}

\vspace{-1em}
\caption{Rewinding to iterations early in training produces subnetworks that outperform the original networks, even when resetting to iteration 0 does not.}
\label{fig:imagenet-accuracy}
\end{figure}

Figure \ref{fig:imagenet-accuracy} illustrates the effect of performing IMP with rewinding across all levels of sparsity.
The blue line shows the result of \emph{one-shot pruning} (pruning all at once after training) at a rewinding iteration just after the aforementioned thresholds.
Resnet-50, Squeezenet, and Inception match the accuracy of the original network when 70\%, 50\%, and 70\% of weights are pruned.
At lower levels of sparsity, these subnetworks slightly outperform the original network.
The weights obtained by rewinding are essential: when randomly reinitialized (dashed blue line) or reset to iteration 0 (orange line), subnetworks lose accuracy when pruned by any amount.

The bottom right plot explores the effect of iterative pruning (training, pruning 20\% of weights at a time, rewinding, and repeating) on Resnet-50.
Rewinding to epoch 6 (the green line) makes it possible to find subnetworks that train to match the original accuracy when just 20\% of weights remain.
Randomly reinitializing (dashed green line) or resetting to iteration 0 (red line) perform equally poorly, only losing accuracy as they are pruned.

\textbf{Discussion.}
On these deeper networks for a more challenging task, IMP finds no evidence in support of \citeauthor{lth}'s hypothesis that equally-capable subnetworks exist at initialization.
However, rewinding to within a few epochs of the start of training makes it possible to find subnetworks with these properties.
Stability continues to offer insight into the value of rewinding: as IMP subnetworks become more stable and a stability gap emerges, they reach higher accuracy.
The central conceit of the lottery ticket hypothesis---that we can prune early in training---continues to apply in this setting;
however, the most productive moment at which to do so is later than strictly at initialization.

\section{Limitations}

Like \citeauthor{lth}'s work, we focus only on image classification.
While we extend their work to include ImageNet, the revised IMP must still train a network one or more times to identify a subnetwork;
we do not propose an efficient way to find these subnetworks at the rewinding iteration.
Our core pruning technique is still
unstructured, magnitude pruning (among many other pruning
techniques, e.g., \citet{prune-activation,data-free-pruning,obs2,pruning-filters,thinet,channel-pruning}).
Unstructured pruning does not necessarily yield networks
that execute more quickly with commodity hardware or libraries; we aim to convey insight on neural
network behavior rather than suggest immediate opportunities to improve performance.

\section{Discussion}

\textbf{Stability.}
Stability to pruning measures a subnetwork's ability to train in isolation to final weights that are close to the values they would have reached had they been trained as part of the original network.
If we consider the trained weights of the original network to be an ideal destination for optimization, then this metric captures a subnetwork's ability to approach this point with fewer parameters.
We conjecture that, by arriving at a similar destination, these subnetworks also reach similar accuracy, potentially explaining why 
increased stability to pruning corresponds to increased accuracy.
\citeauthor{lth}'s winning tickets are substantially more stable to pruning than random subnetworks, shedding light on a possible mechanism behind the lottery ticket hypothesis.
As a follow-up, one might explore whether subnetworks that are more stable to pruning are also more likely to reach the same basin of attraction as the original network \citep{kolter}.

We find a similar relationship between stability to data order and accuracy.
Stability to data order measures a subnetwork's ability to reach similar final weights in spite of training with different mini-batch sequences---the gradient noise intrinsic to SGD.
This metric is valuable because it provides a way to measure subnetwork stability without reference to the original network.
We believe that stability to pruning and data order work hand-in-hand: subnetworks that are robust to data order are better able to reach destinations consistent with the original network in the face of SGD noise.

\textbf{Pruning early.}
We aim to characterize the opportunity to prune early in training.
Doing so could make it possible to reduce the cost of training networks by substantially reducing parameter-counts for most or all of training.
This is an active area of research in its own right, including pruning before training \citep{snip}, pruning throughout training \citep{zhu2017prune, gale, lym2019prunetrain, variational-sparsifies, l0-reg, exploring}, and maintaining a sparse network with dynamic structure \citep{deepr, mostafa2018dynamic, mocanu2018scalable}.
However, to the best of our knowledge, our work is the first to show that it is possible to prune (1) so early in training (2) to such extreme levels of sparsity (3) on such large-scale tasks.

Specifically, we find that, for many networks, there is an iteration early in training after which pruning can result in subnetworks with far higher accuracy than when pruning at initialization.
Our results with IMP expand the range of known opportunities to prune early in training that---if exploited---could reduce the cost of training.
With better techniques, we expect this range could be expanded even further because our results are restricted by IMP's limitations. Namely, 
it is possible that there are equally-capable subnetworks present at initialization, but IMP is unable to find them.

Stability offers a new perspective for developing new early pruning methods.
One could exploit stability information for pruning, or develop new techniques to maintain stability under pruning.
Under our hypothesized connection between stability and accuracy, such methods could make it possible to find accurate subnetworks early in training.

\section{Conclusion}

The lottery ticket hypothesis hints at future techniques that
identify small, trainable subnetworks capable of matching the accuracy of the larger networks we typically train.
To date, this and other related research have focused on
compressing neural networks before training.
In this work, we
find that other moments early in the training process may present better opportunities for this class of techniques.
In doing so, we shed new light on the lottery ticket hypothesis and its manifestation in deeper networks through the lens of stability.

\bibliography{local}


\begin{thebibliography}{27}

%%% the Web address.

\ifx \showCODEN    \undefined \def \showCODEN     #1{\unskip}     \fi
\ifx \showDOI      \undefined \def \showDOI       #1{#1}\fi
\ifx \showISBNx    \undefined \def \showISBNx     #1{\unskip}     \fi
\ifx \showISBNxiii \undefined \def \showISBNxiii  #1{\unskip}     \fi
\ifx \showISSN     \undefined \def \showISSN      #1{\unskip}     \fi
\ifx \showLCCN     \undefined \def \showLCCN      #1{\unskip}     \fi
\ifx \shownote     \undefined \def \shownote      #1{#1}          \fi
\ifx \showarticletitle \undefined \def \showarticletitle #1{#1}   \fi
\ifx \showURL      \undefined \def \showURL       {\relax}        \fi
% invisible to TeX
\providecommand\bibfield[2]{#2}
\providecommand\bibinfo[2]{#2}
\providecommand\natexlab[1]{#1}
\providecommand\showeprint[2][]{arXiv:#2}

\bibitem[\protect\citeauthoryear{Bellec, Kappel, Maass, and Legenstein}{Bellec
  et~al\mbox{.}}{2018}]%
        {deepr}
\bibfield{author}{\bibinfo{person}{Guillaume Bellec}, \bibinfo{person}{David
  Kappel}, \bibinfo{person}{Wolfgang Maass}, {and} \bibinfo{person}{Robert
  Legenstein}.} \bibinfo{year}{2018}\natexlab{}.
\newblock \showarticletitle{Deep Rewiring: Training very sparse deep networks}.
\newblock \bibinfo{journal}{\emph{Proceedings of ICLR}} (\bibinfo{year}{2018}).
\newblock


\bibitem[\protect\citeauthoryear{Dong, Chen, and Pan}{Dong
  et~al\mbox{.}}{2017}]%
        {obs2}
\bibfield{author}{\bibinfo{person}{Xin Dong}, \bibinfo{person}{Shangyu Chen},
  {and} \bibinfo{person}{Sinno Pan}.} \bibinfo{year}{2017}\natexlab{}.
\newblock \showarticletitle{Learning to prune deep neural networks via
  layer-wise optimal brain surgeon}. In \bibinfo{booktitle}{\emph{Advances in
  Neural Information Processing Systems}}. \bibinfo{pages}{4860--4874}.
\newblock


\bibitem[\protect\citeauthoryear{Frankle and Carbin}{Frankle and
  Carbin}{2019}]%
        {lth}
\bibfield{author}{\bibinfo{person}{Jonathan Frankle} {and}
  \bibinfo{person}{Michael Carbin}.} \bibinfo{year}{2019}\natexlab{}.
\newblock \showarticletitle{The Lottery Ticket Hypothesis: Finding Sparse,
  Trainable Neural Networks}. In \bibinfo{booktitle}{\emph{Int. Conf.
  Represent. Learn.}}
\newblock
\showeprint[arxiv]{1803.03635}


\bibitem[\protect\citeauthoryear{Frankle, Dziugaite, Roy, and Carbin}{Frankle
  et~al\mbox{.}}{2020}]%
        {frankle2019lottery}
\bibfield{author}{\bibinfo{person}{Jonathan Frankle},
  \bibinfo{person}{Gintare~Karolina Dziugaite}, \bibinfo{person}{Daniel~M Roy},
  {and} \bibinfo{person}{Michael Carbin}.} \bibinfo{year}{2020}\natexlab{}.
\newblock \showarticletitle{Linear Mode Connectivity and the Lottery Ticket
  Hypothesis}. In \bibinfo{booktitle}{\emph{International Conference on Machine
  Learning}}.
\newblock


\bibitem[\protect\citeauthoryear{Gale, Elsen, and Hooker}{Gale
  et~al\mbox{.}}{2019}]%
        {gale}
\bibfield{author}{\bibinfo{person}{Trevor Gale}, \bibinfo{person}{Erich Elsen},
  {and} \bibinfo{person}{Sara Hooker}.} \bibinfo{year}{2019}\natexlab{}.
\newblock \showarticletitle{The State of Sparsity in Deep Neural Networks}.
\newblock \bibinfo{journal}{\emph{arXiv preprint arXiv:1902.09574}}
  (\bibinfo{year}{2019}).
\newblock


\bibitem[\protect\citeauthoryear{Google}{Google}{2018}]%
        {tpu-implementation}
\bibfield{author}{\bibinfo{person}{Google}.} \bibinfo{year}{2018}\natexlab{}.
\newblock \bibinfo{title}{Networks for Imagenet on TPUs}.
\newblock   (\bibinfo{year}{2018}).
\newblock
\urldef\tempurl%
\url{https://github.com/tensorflow/tpu/tree/master/models/}
\showURL{%
\tempurl}


\bibitem[\protect\citeauthoryear{Han, Pool, Tran, and Dally}{Han
  et~al\mbox{.}}{2015}]%
        {han-pruning}
\bibfield{author}{\bibinfo{person}{Song Han}, \bibinfo{person}{Jeff Pool},
  \bibinfo{person}{John Tran}, {and} \bibinfo{person}{William Dally}.}
  \bibinfo{year}{2015}\natexlab{}.
\newblock \showarticletitle{Learning both weights and connections for efficient
  neural network}. In \bibinfo{booktitle}{\emph{Advances in neural information
  processing systems}}. \bibinfo{pages}{1135--1143}.
\newblock


\bibitem[\protect\citeauthoryear{He, Zhang, Ren, and Sun}{He
  et~al\mbox{.}}{2016}]%
        {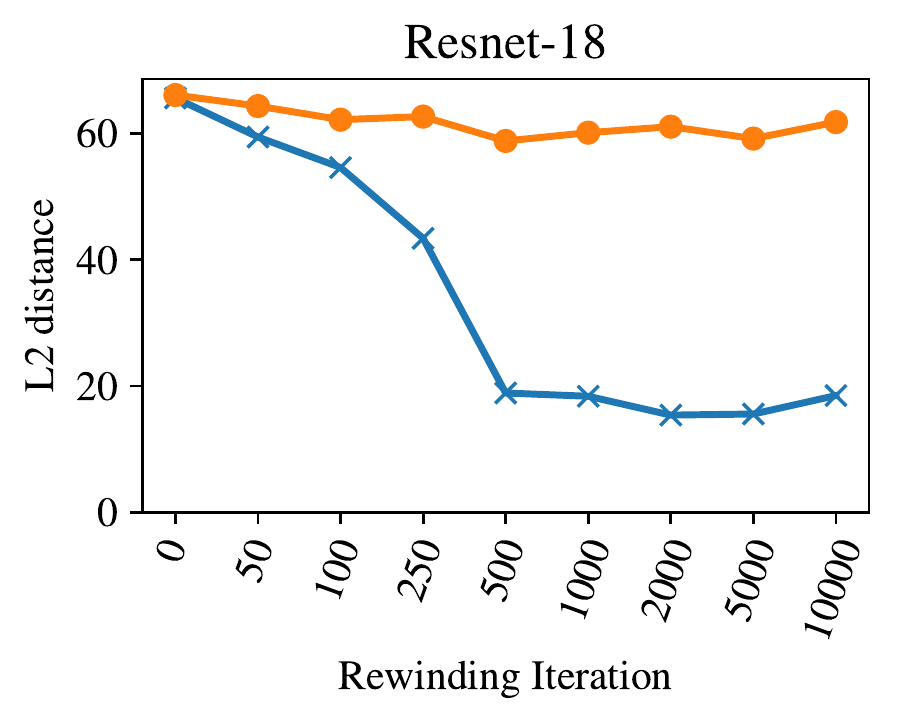}
\bibfield{author}{\bibinfo{person}{Kaiming He}, \bibinfo{person}{Xiangyu
  Zhang}, \bibinfo{person}{Shaoqing Ren}, {and} \bibinfo{person}{Jian Sun}.}
  \bibinfo{year}{2016}\natexlab{}.
\newblock \showarticletitle{Deep residual learning for image recognition}. In
  \bibinfo{booktitle}{\emph{Proceedings of the IEEE Conference on Computer
  Vision and Pattern Recognition}}. \bibinfo{pages}{770--778}.
\newblock


\bibitem[\protect\citeauthoryear{He, Zhang, and Sun}{He et~al\mbox{.}}{2017}]%
        {channel-pruning}
\bibfield{author}{\bibinfo{person}{Yihui He}, \bibinfo{person}{Xiangyu Zhang},
  {and} \bibinfo{person}{Jian Sun}.} \bibinfo{year}{2017}\natexlab{}.
\newblock \showarticletitle{Channel pruning for accelerating very deep neural
  networks}. In \bibinfo{booktitle}{\emph{International Conference on Computer
  Vision (ICCV)}}, Vol.~\bibinfo{volume}{2}. \bibinfo{pages}{6}.
\newblock


\bibitem[\protect\citeauthoryear{Hu, Peng, Tai, and Tang}{Hu
  et~al\mbox{.}}{2016}]%
        {prune-activation}
\bibfield{author}{\bibinfo{person}{Hengyuan Hu}, \bibinfo{person}{Rui Peng},
  \bibinfo{person}{Yu-Wing Tai}, {and} \bibinfo{person}{Chi-Keung Tang}.}
  \bibinfo{year}{2016}\natexlab{}.
\newblock \showarticletitle{Network trimming: A data-driven neuron pruning
  approach towards efficient deep architectures}.
\newblock \bibinfo{journal}{\emph{arXiv preprint arXiv:1607.03250}}
  (\bibinfo{year}{2016}).
\newblock


\bibitem[\protect\citeauthoryear{Iandola, Han, Moskewicz, Ashraf, Dally, and
  Keutzer}{Iandola et~al\mbox{.}}{2016}]%
        {squeezenet}
\bibfield{author}{\bibinfo{person}{Forrest~N Iandola}, \bibinfo{person}{Song
  Han}, \bibinfo{person}{Matthew~W Moskewicz}, \bibinfo{person}{Khalid Ashraf},
  \bibinfo{person}{William~J Dally}, {and} \bibinfo{person}{Kurt Keutzer}.}
  \bibinfo{year}{2016}\natexlab{}.
\newblock \showarticletitle{SqueezeNet: AlexNet-level accuracy with 50x fewer
  parameters and< 0.5 MB model size}.
\newblock \bibinfo{journal}{\emph{arXiv preprint arXiv:1602.07360}}
  (\bibinfo{year}{2016}).
\newblock


\bibitem[\protect\citeauthoryear{LeCun, Denker, and Solla}{LeCun
  et~al\mbox{.}}{1990}]%
        {brain-damage}
\bibfield{author}{\bibinfo{person}{Yann LeCun}, \bibinfo{person}{John~S
  Denker}, {and} \bibinfo{person}{Sara~A Solla}.}
  \bibinfo{year}{1990}\natexlab{}.
\newblock \showarticletitle{Optimal brain damage}. In
  \bibinfo{booktitle}{\emph{Advances in neural information processing
  systems}}. \bibinfo{pages}{598--605}.
\newblock


\bibitem[\protect\citeauthoryear{Lee, Ajanthan, and Torr}{Lee
  et~al\mbox{.}}{2019}]%
        {snip}
\bibfield{author}{\bibinfo{person}{Namhoon Lee},
  \bibinfo{person}{Thalaiyasingam Ajanthan}, {and} \bibinfo{person}{Philip~HS
  Torr}.} \bibinfo{year}{2019}\natexlab{}.
\newblock \showarticletitle{SNIP: Single-shot Network Pruning based on
  Connection Sensitivity}.
\newblock  (\bibinfo{year}{2019}).
\newblock


\bibitem[\protect\citeauthoryear{Li, Kadav, Durdanovic, Samet, and Graf}{Li
  et~al\mbox{.}}{2016}]%
        {pruning-filters}
\bibfield{author}{\bibinfo{person}{Hao Li}, \bibinfo{person}{Asim Kadav},
  \bibinfo{person}{Igor Durdanovic}, \bibinfo{person}{Hanan Samet}, {and}
  \bibinfo{person}{Hans~Peter Graf}.} \bibinfo{year}{2016}\natexlab{}.
\newblock \showarticletitle{Pruning filters for efficient convnets}.
\newblock \bibinfo{journal}{\emph{arXiv preprint arXiv:1608.08710}}
  (\bibinfo{year}{2016}).
\newblock


\bibitem[\protect\citeauthoryear{Liu, Sun, Zhou, Huang, and Darrell}{Liu
  et~al\mbox{.}}{2019}]%
        {rethinking-pruning}
\bibfield{author}{\bibinfo{person}{Zhuang Liu}, \bibinfo{person}{Mingjie Sun},
  \bibinfo{person}{Tinghui Zhou}, \bibinfo{person}{Gao Huang}, {and}
  \bibinfo{person}{Trevor Darrell}.} \bibinfo{year}{2019}\natexlab{}.
\newblock \showarticletitle{Rethinking the Value of Network Pruning}. In
  \bibinfo{booktitle}{\emph{International Conference on Learning
  Representations}}.
\newblock
\urldef\tempurl%
\url{https://openreview.net/forum? id=rJlnB3C5Ym}
\showURL{%
\tempurl}


\bibitem[\protect\citeauthoryear{Louizos, Welling, and Kingma}{Louizos
  et~al\mbox{.}}{2018}]%
        {l0-reg}
\bibfield{author}{\bibinfo{person}{Christos Louizos}, \bibinfo{person}{Max
  Welling}, {and} \bibinfo{person}{Diederik~P Kingma}.}
  \bibinfo{year}{2018}\natexlab{}.
\newblock \showarticletitle{Learning Sparse Neural Networks through $ L\_0 $
  Regularization}.
\newblock \bibinfo{journal}{\emph{Proceedings of ICLR}} (\bibinfo{year}{2018}).
\newblock


\bibitem[\protect\citeauthoryear{Luo, Wu, and Lin}{Luo et~al\mbox{.}}{2017}]%
        {thinet}
\bibfield{author}{\bibinfo{person}{Jian-Hao Luo}, \bibinfo{person}{Jianxin Wu},
  {and} \bibinfo{person}{Weiyao Lin}.} \bibinfo{year}{2017}\natexlab{}.
\newblock \showarticletitle{Thinet: A filter level pruning method for deep
  neural network compression}.
\newblock \bibinfo{journal}{\emph{arXiv preprint arXiv:1707.06342}}
  (\bibinfo{year}{2017}).
\newblock


\bibitem[\protect\citeauthoryear{Lym, Choukse, Zangeneh, Wen, Erez, and
  Shanghavi}{Lym et~al\mbox{.}}{2019}]%
        {lym2019prunetrain}
\bibfield{author}{\bibinfo{person}{Sangkug Lym}, \bibinfo{person}{Esha
  Choukse}, \bibinfo{person}{Siavash Zangeneh}, \bibinfo{person}{Wei Wen},
  \bibinfo{person}{Mattan Erez}, {and} \bibinfo{person}{Sujay Shanghavi}.}
  \bibinfo{year}{2019}\natexlab{}.
\newblock \showarticletitle{PruneTrain: Gradual Structured Pruning from Scratch
  for Faster Neural Network Training}.
\newblock \bibinfo{journal}{\emph{arXiv preprint arXiv:1901.09290}}
  (\bibinfo{year}{2019}).
\newblock


\bibitem[\protect\citeauthoryear{Mocanu, Mocanu, Stone, Nguyen, Gibescu, and
  Liotta}{Mocanu et~al\mbox{.}}{2018}]%
        {mocanu2018scalable}
\bibfield{author}{\bibinfo{person}{Decebal~Constantin Mocanu},
  \bibinfo{person}{Elena Mocanu}, \bibinfo{person}{Peter Stone},
  \bibinfo{person}{Phuong~H Nguyen}, \bibinfo{person}{Madeleine Gibescu}, {and}
  \bibinfo{person}{Antonio Liotta}.} \bibinfo{year}{2018}\natexlab{}.
\newblock \showarticletitle{Scalable training of artificial neural networks
  with adaptive sparse connectivity inspired by network science}.
\newblock \bibinfo{journal}{\emph{Nature communications}} \bibinfo{volume}{9},
  \bibinfo{number}{1} (\bibinfo{year}{2018}), \bibinfo{pages}{2383}.
\newblock


\bibitem[\protect\citeauthoryear{Molchanov, Ashukha, and Vetrov}{Molchanov
  et~al\mbox{.}}{2017}]%
        {variational-sparsifies}
\bibfield{author}{\bibinfo{person}{Dmitry Molchanov}, \bibinfo{person}{Arsenii
  Ashukha}, {and} \bibinfo{person}{Dmitry Vetrov}.}
  \bibinfo{year}{2017}\natexlab{}.
\newblock \showarticletitle{Variational dropout sparsifies deep neural
  networks}.
\newblock \bibinfo{journal}{\emph{arXiv preprint arXiv:1701.05369}}
  (\bibinfo{year}{2017}).
\newblock


\bibitem[\protect\citeauthoryear{Mostafa and Wang}{Mostafa and Wang}{2018}]%
        {mostafa2018dynamic}
\bibfield{author}{\bibinfo{person}{Hesham Mostafa} {and} \bibinfo{person}{Xin
  Wang}.} \bibinfo{year}{2018}\natexlab{}.
\newblock \showarticletitle{Dynamic parameter reallocation improves
  trainability of deep convolutional networks}.
\newblock  (\bibinfo{year}{2018}).
\newblock


\bibitem[\protect\citeauthoryear{Nagarajan and Kolter}{Nagarajan and
  Kolter}{2019}]%
        {kolter}
\bibfield{author}{\bibinfo{person}{Vaishnavh Nagarajan} {and}
  \bibinfo{person}{J.~Zico Kolter}.} \bibinfo{year}{2019}\natexlab{}.
\newblock \showarticletitle{Uniform convergence may be unable to explain
  generalization in deep learning}.
\newblock \bibinfo{journal}{\emph{CoRR}}  \bibinfo{volume}{abs/1902.04742}
  (\bibinfo{year}{2019}).
\newblock
\showeprint[arxiv]{1902.04742}
\urldef\tempurl%
\url{http://arxiv.org/abs/1902.04742}
\showURL{%
\tempurl}


\bibitem[\protect\citeauthoryear{Narang, Elsen, Diamos, and Sengupta}{Narang
  et~al\mbox{.}}{2017}]%
        {exploring}
\bibfield{author}{\bibinfo{person}{Sharan Narang}, \bibinfo{person}{Erich
  Elsen}, \bibinfo{person}{Gregory Diamos}, {and} \bibinfo{person}{Shubho
  Sengupta}.} \bibinfo{year}{2017}\natexlab{}.
\newblock \showarticletitle{Exploring sparsity in recurrent neural networks}.
\newblock \bibinfo{journal}{\emph{Proceedings of ICLR}} (\bibinfo{year}{2017}).
\newblock


\bibitem[\protect\citeauthoryear{Russakovsky, Deng, Su, Krause, Satheesh, Ma,
  Huang, Karpathy, Khosla, Bernstein, et~al\mbox{.}}{Russakovsky
  et~al\mbox{.}}{2015}]%
        {imagenet}
\bibfield{author}{\bibinfo{person}{Olga Russakovsky}, \bibinfo{person}{Jia
  Deng}, \bibinfo{person}{Hao Su}, \bibinfo{person}{Jonathan Krause},
  \bibinfo{person}{Sanjeev Satheesh}, \bibinfo{person}{Sean Ma},
  \bibinfo{person}{Zhiheng Huang}, \bibinfo{person}{Andrej Karpathy},
  \bibinfo{person}{Aditya Khosla}, \bibinfo{person}{Michael Bernstein},
  {et~al\mbox{.}}} \bibinfo{year}{2015}\natexlab{}.
\newblock \showarticletitle{Imagenet large scale visual recognition challenge}.
\newblock \bibinfo{journal}{\emph{International Journal of Computer Vision}}
  \bibinfo{volume}{115}, \bibinfo{number}{3} (\bibinfo{year}{2015}),
  \bibinfo{pages}{211--252}.
\newblock


\bibitem[\protect\citeauthoryear{Srinivas and Babu}{Srinivas and Babu}{2015}]%
        {data-free-pruning}
\bibfield{author}{\bibinfo{person}{Suraj Srinivas} {and}
  \bibinfo{person}{R~Venkatesh Babu}.} \bibinfo{year}{2015}\natexlab{}.
\newblock \showarticletitle{Data-free parameter pruning for deep neural
  networks}.
\newblock \bibinfo{journal}{\emph{arXiv preprint arXiv:1507.06149}}
  (\bibinfo{year}{2015}).
\newblock


\bibitem[\protect\citeauthoryear{Szegedy, Vanhoucke, Ioffe, Shlens, and
  Wojna}{Szegedy et~al\mbox{.}}{2016}]%
        {inception2}
\bibfield{author}{\bibinfo{person}{Christian Szegedy}, \bibinfo{person}{Vincent
  Vanhoucke}, \bibinfo{person}{Sergey Ioffe}, \bibinfo{person}{Jon Shlens},
  {and} \bibinfo{person}{Zbigniew Wojna}.} \bibinfo{year}{2016}\natexlab{}.
\newblock \showarticletitle{Rethinking the inception architecture for computer
  vision}. In \bibinfo{booktitle}{\emph{Proceedings of the IEEE conference on
  computer vision and pattern recognition}}. \bibinfo{pages}{2818--2826}.
\newblock


\bibitem[\protect\citeauthoryear{Zhu and Gupta}{Zhu and Gupta}{2017}]%
        {zhu2017prune}
\bibfield{author}{\bibinfo{person}{Michael Zhu} {and} \bibinfo{person}{Suyog
  Gupta}.} \bibinfo{year}{2017}\natexlab{}.
\newblock \showarticletitle{To prune, or not to prune: exploring the efficacy
  of pruning for model compression}.
\newblock \bibinfo{journal}{\emph{arXiv preprint arXiv:1710.01878}}
  (\bibinfo{year}{2017}).
\newblock


\end{thebibliography}
\bibliographystyle{ACM-Reference-Format}

\newpage
\begin{appendix}

%%%%%%%%%%%%%%%%%%%%%%%%%%%%%%%%%%%%%%%%%%%%%%%%%%%%%%%%%%%%%%%%%%%%%%%%%%%%%%%%%%
% APPENDICES
%%%%%%%%%%%%%%%%%%%%%%%%%%%%%%%%%%%%%%%%%%%%%%%%%%%%%%%%%%%%%%%%%%%%%%%%%%%%%%%%%%
\appendix

\section{Comparison to ``Rethinking the Value of Pruning'' and ``SNIP''}
\label{app:comp}

In this appendix, we compare the performance of subnetworks found by ``The Lottery Ticket Hypothesis'' \citep{lth}, ``Rethinking the Value of Pruning'' \citep{rethinking-pruning}, and ``SNIP'' \citep{snip}.
All three papers have different perspectives on the prospect of pruning early in training.

\citeauthor{lth} argue that sparse, trainable networks exist at initialization time within the networks that we typically train.
They find these networks using IMP (\ref{fig:algorithm1} with $k = 0$) to find these subnetworks: they train the original network, prune it, and reset each surviving connection's weight back to its initial value from before training.
They argue that the original initializations are essential for achieving this performance and that randomly reinitializing substantially degrades performance.

\citeauthor{rethinking-pruning} argue that the sparse networks that result from pruning can be trained from the start and that the original initializations do not matter.
Instead, they find that the networks that result from pruning can be trained with a random initialization to the same performance as the original network.

\citeauthor{snip} propose a method for pruning early in training called SNIP.
SNIP considers the \emph{sensitivity} of the loss to each weight (based on one mini-batch of data) and removes those weights to which the loss is least sensitive.
Sensitivity is measured by multiplying each weight $w$ by a virtual
parameter $c = 1$ and computing $\frac{\partial L}{\partial c}$.
The authors find that SNIP can prune neural networks before they have been trained, and that these networks can be randomly reinitialized without harm to the eventual accuracy.

There are three points of contention between the papers:
\begin{enumerate}
\item Is the original initialization important? \citeauthor{lth} argue that it is essential, but \citeauthor{rethinking-pruning} and \citeauthor{snip} discard it without any impact on their results.
\item At what level of sparsity are the authors measuring? For example, \citeauthor{rethinking-pruning} and \citeauthor{snip} consider VGG-19 when pruned by up to 95\%, while \citeauthor{lth} prune by upwards of 99.5\%.
\item How efficient is the pruning method? \citeauthor{lth} must train the same network a dozen or more times, while \citeauthor{rethinking-pruning} must train the network once and \citeauthor{snip} need to only look at a single mini-batch of data.
\end{enumerate}

\begin{figure}
  \centering
    \vspace{-.5em}
    \includegraphics[width=0.5\textwidth]{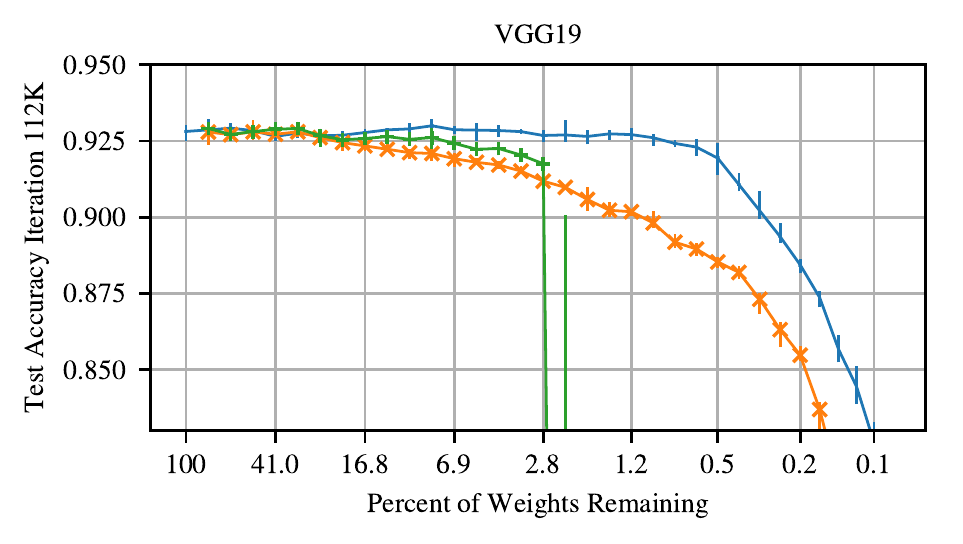}%
    \includegraphics[width=0.5\textwidth]{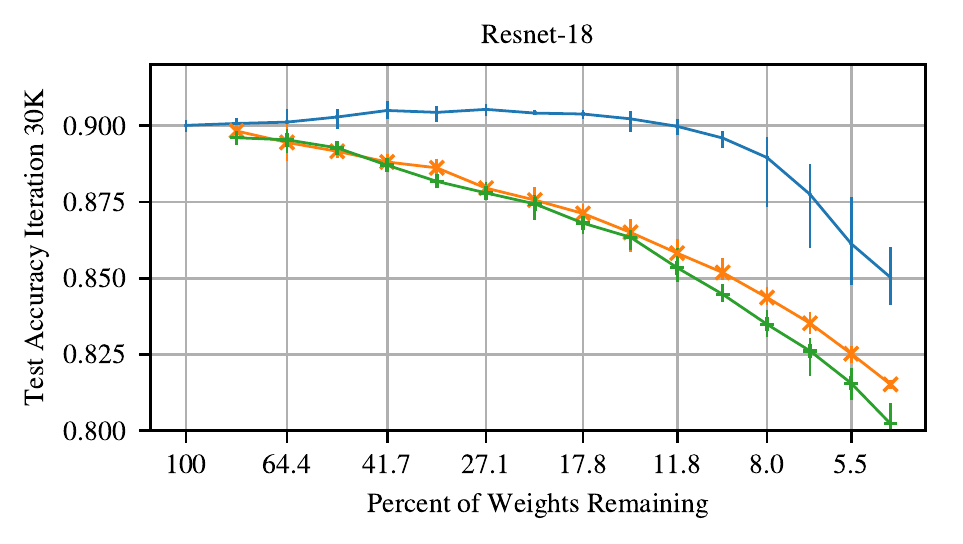}%
    \vspace{-1em}
    \includegraphics[width=0.6\textwidth]{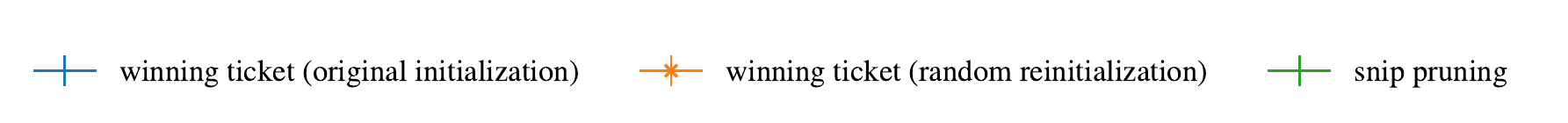}%
    \vspace{-1em}
    \caption{The accuracy achieved by VGG19 (left) and Resnet-18 (right) on CIFAR10
    when pruned to the specified size using iterative pruning and SNIP. Networks are trained with warmup and
    the learning rate hyperparameters used by \citeauthor{lth}.}
    \label{fig:lth-snip}
\end{figure}

Figure~\ref{fig:lth-snip} compares these three methods on VGG-19 (warmup) and Resnet-18 (warmup) from Figure \ref{fig:small-networks}.
These particular hyperparameters were chosen because IMP does not find winning tickets without warmup, which would render this comparison less informative.
The plots include the accuracy of randomly reinitialized winning tickets (orange), winning tickets with the original initialization (blue), and subnetworks found by SNIP (green).
The results for VGG19 (left) support the findings of \citeauthor{rethinking-pruning}
that pruned, randomly reinitialized networks
can match the accuracy of the original network: VGG19 can do so when pruned by up to 80\%.
However, beyond this point, the accuracy of the randomly
reinitialized networks declines steadily. In contrast, winning tickets with the original initialization match
the accuracy of the original network when pruned by up to 99\%. For Resnet-18 (right), which has 75x
fewer parameters, the randomly reinitialized networks lose accuracy much sooner.

SNIP results in a promising improvement over random reinitialization on VGG19, however there is still a performance
gap between SNIP and the winning tickets---an opportunity to further improve the performance of pruning before training.

We return to our original three questions:
Up to a certain level of sparsity, the original initialization is not important. Subnetworks that are randomly reinitialized can still train to full accuracy. Beyond this point, however, the original initialization is necessary in order to maintain high accuracy. \citeauthor{rethinking-pruning} and \citeauthor{snip} operate in this first regime, where initialization is less important; \citeauthor{lth} operate in the second regime.
However, finding networks that learn effectively at these extreme levels of sparsity is very expensive: \citeauthor{lth} must train the same network many times in order to do so.
\citeauthor{snip} offer a promising direction for efficiently finding such subnetworks, taking a step toward realizing the opportunity described by \citeauthor{lth} and this paper.

\section{Angle Measurements (Stability)}
\label{app:angle}

\begin{figure}
\centering
\scriptsize
\begin{tabular}{l@{\ \ }|@{\ \ }c@{\ \ } | c@{\ \ \ \ }c@{\ \ \ \ }c | c@{\ \ \ \ }c@{\ \ \ \ }c | c@{\ \ \ \ }c }
\toprule
Network & Sparsity & \multicolumn{3}{c|}{Data Order Stability (Angle)} & \multicolumn{3}{c|}{Pruning Stability (Angle)} & \multicolumn{2}{c}{Accuracy} \\
& & IMP & Random & Comp & IMP & Random & Comp & IMP & Random \\  \midrule
Lenet  & 10.7\% & 22.7 $\pm$ 0.5$\degree$ & 49.2 $\pm$ 3.0$\degree$ & 2.2x & 53.1 $\pm$ 0.8$\degree$ & 83.6 $\pm$ 1.1$\degree$ & 1.6x & 98.3 $\pm$ 0.1 & 97.5 $\pm$ 0.3 \\ \midrule
Resnet-18 (standard)&  & 87.7 $\pm$ 1.2$\degree$ & 88.0 $\pm$ 1.4$\degree$ & 1.0x & 87.7 $\pm$ 1.2 $\degree$ & 88.0 $\pm$ 1.4$\degree$ & 1.0x & 87.7 $\pm$ 0.4 & 87.7 $\pm$ 0.5 \\
Resnet-18 (low) & 16.7\% & 16.1 $\pm$ 2.7$\degree$ & 75.8 $\pm$ 1.2$\degree$ & 4.7x & 50.4 $\pm$ 4.8$\degree$ &  77.1 $\pm$ 0.4$\degree$ & 1.5x &  89.1 $\pm$ 0.4 & 86.1 $\pm$ 0.6 \\
Resnet-18 (warmup)& & 17.0 $\pm$ 0.2$\degree$ & 69.0 $\pm$ 3.8$\degree$ & 4.7x  & 49.9 $\pm$ 2.3$\degree$ & 79.0 $\pm$ 0.5$\degree$ & 1.6x &  90.3 $\pm$ 0.4 & 86.8 $\pm$ 0.5 \\ \midrule
VGG-19 (standard) & & 88.5 $\pm$ 0.4$\degree$ & 88.3 $\pm$ 0.5$\degree$ & 1.0x & 87.8 $\pm$ 0.2$\degree$ & 88.1 $\pm$ 0.2$\degree$ & 1.0x & 90.0 $\pm$ 0.3 & 90.2 $\pm$ 0.5 \\
VGG-19 (low) & 2.2\% &39.9 $\pm$ 2.1$\degree$ & 84.9 $\pm$ 0.6$\degree$ & 2.1x & 54.0 $\pm$ 0.4$\degree$ & 79.0 $\pm$ 0.2$\degree$ & 1.5x& 91.0 $\pm$ 0.3 & 88.0 $\pm$ 0.3 \\
VGG-19 (warmup) & & 39.9 $\pm$ 2.1$\degree$ & 84.6 $\pm$ 0.4$\degree$ & 2.1x &  71.7 $\pm$ 0.3$\degree$ & 84.8 $\pm$ 0.2$\degree$ & 1.2x & 92.4 $\pm$ 0.2 & 90.3 $\pm$ 0.3 \\
\bottomrule
\end{tabular}
\caption{The average stability (as measured in angle) of subnetworks obtained by IMP and by randomly pruning. Errors are the minimum or maximum across 18 samples (data order) and 3 samples (pruning and accuracy).}
\label{fig:stability-at-zero-angle}
\end{figure}

\begin{figure}
\includegraphics[width=0.25\textwidth]{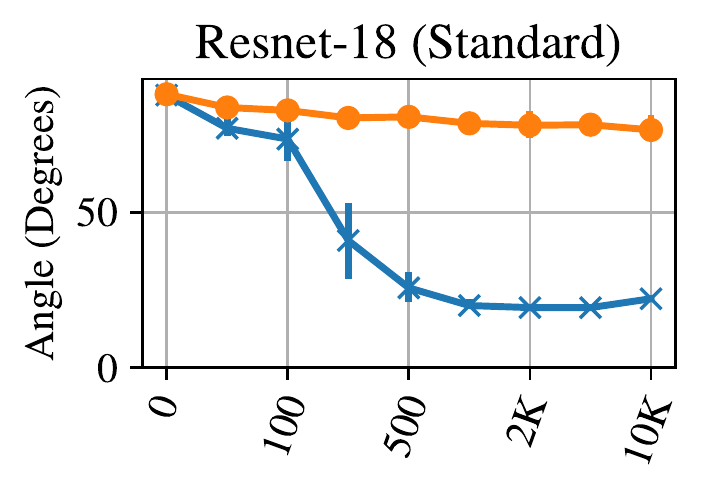}%
\includegraphics[width=0.25\textwidth]{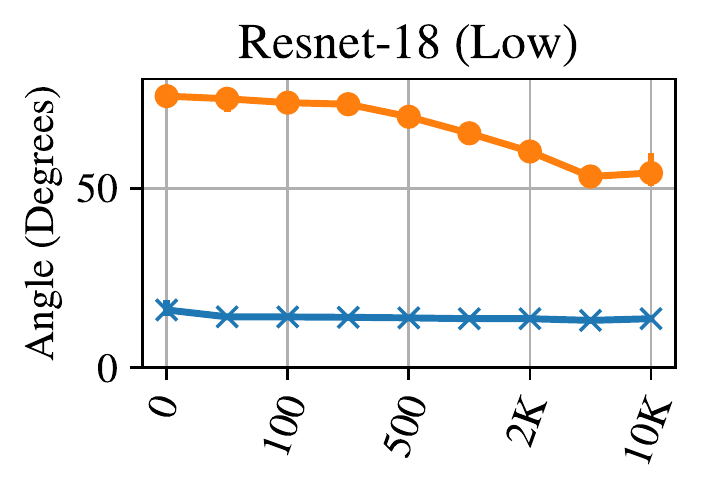}%
\includegraphics[width=0.25\textwidth]{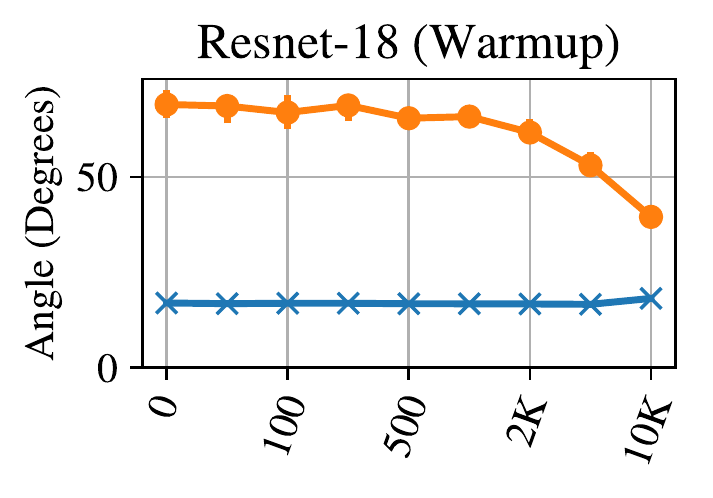}%
\includegraphics[width=0.25\textwidth]{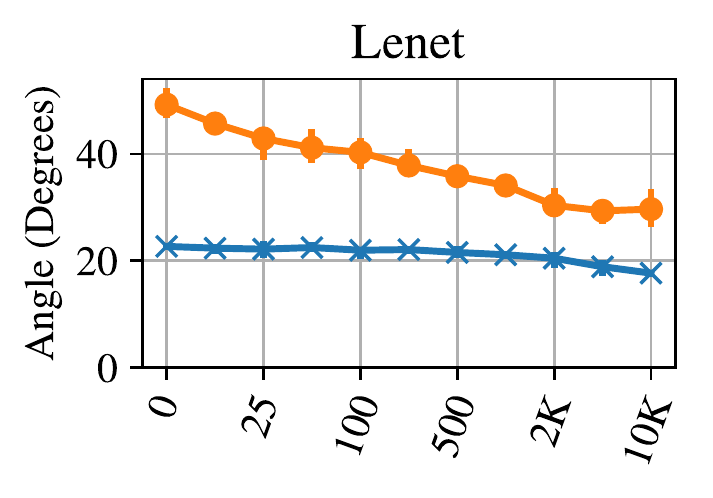}

\vspace{-.6em}%
\includegraphics[width=0.25\textwidth]{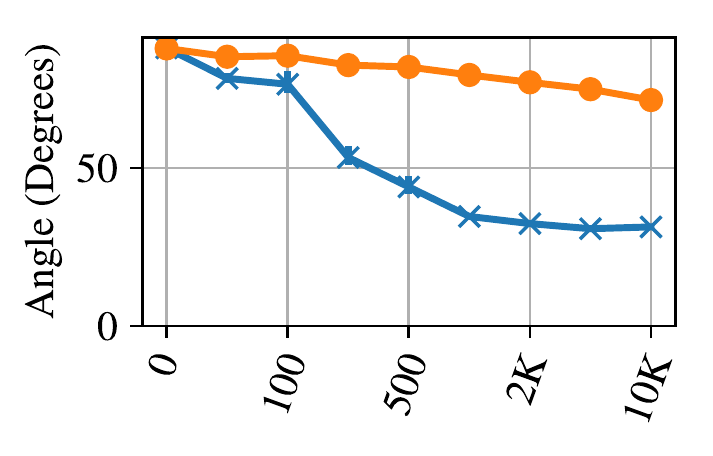}%
\includegraphics[width=0.25\textwidth]{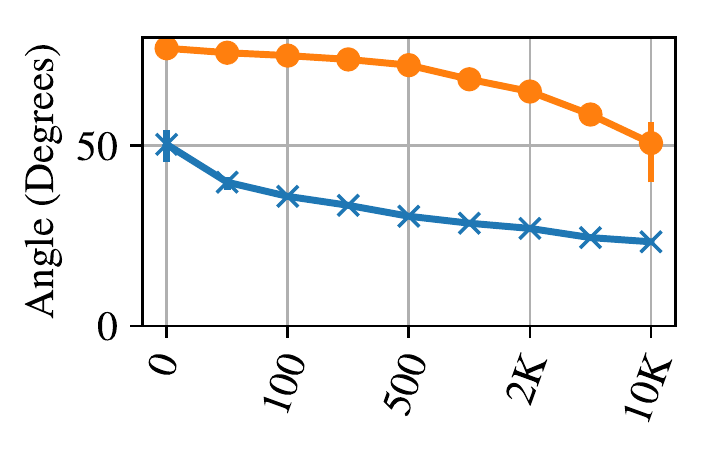}%
\includegraphics[width=0.25\textwidth]{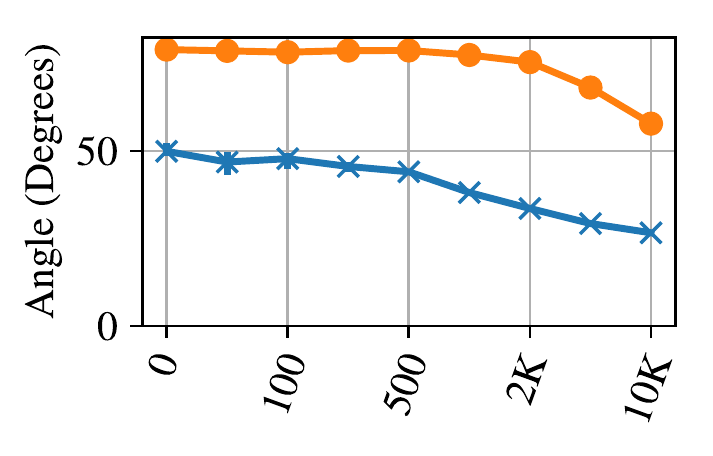}%
\includegraphics[width=0.25\textwidth]{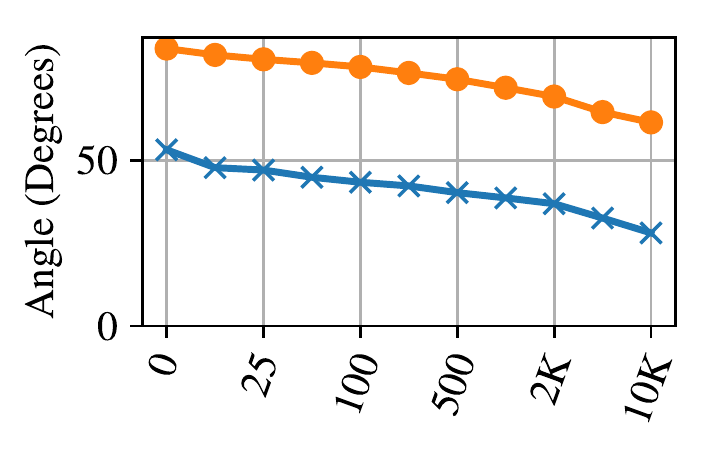}

\vspace{-.6em}%
\includegraphics[width=0.25\textwidth]{figures/late/resnet18-accuracy-dataorder}%
\includegraphics[width=0.25\textwidth]{figures/late/resnet-low-accuracy-dataorder}%
\includegraphics[width=0.25\textwidth]{figures/late/resnet-warmup-accuracy-dataorder}%
\includegraphics[width=0.25\textwidth]{figures/late/lenet-accuracy-dataorder}

\vspace{-.6em}%
\includegraphics[width=0.25\textwidth]{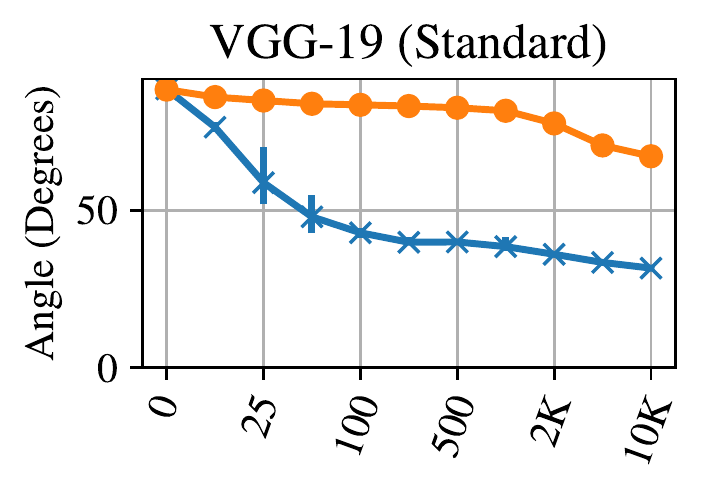}%
\includegraphics[width=0.25\textwidth]{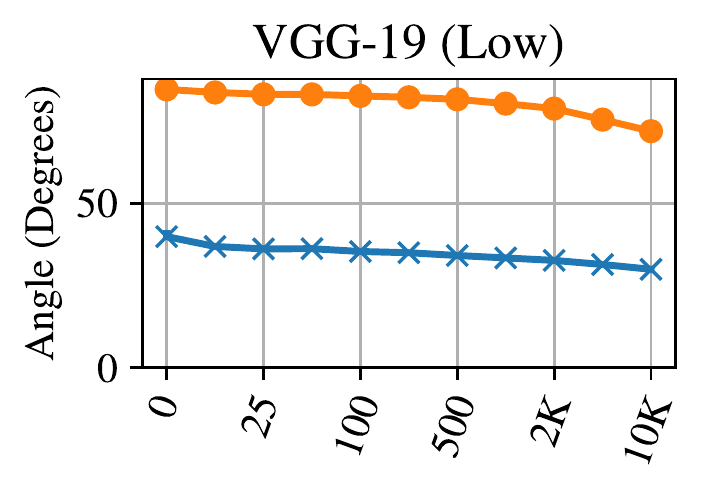}%
\includegraphics[width=0.25\textwidth]{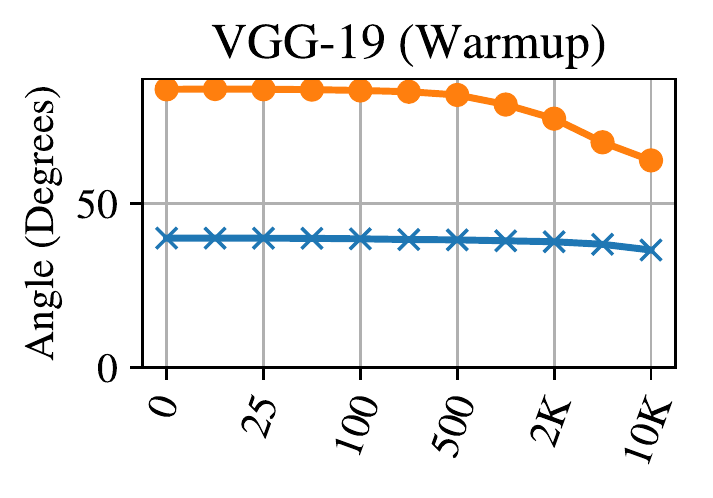}%
\includegraphics[width=.2\textwidth]{figures/late/lenet-accuracy-dataorder-legend}\hspace{.05\textwidth}

\vspace{-.6em}%
\includegraphics[width=0.25\textwidth]{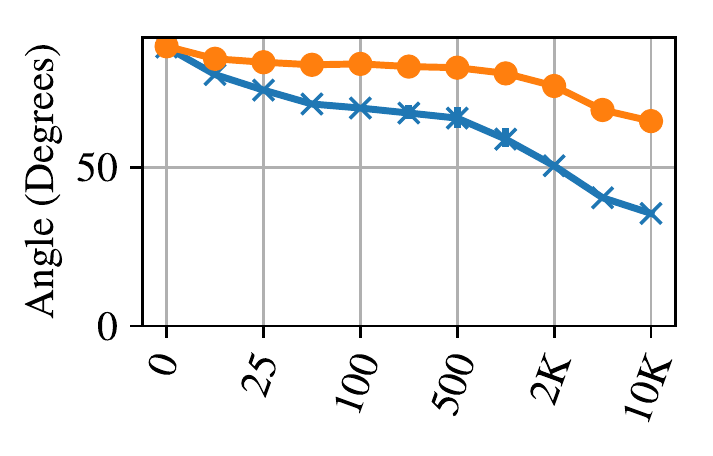}%
\includegraphics[width=0.25\textwidth]{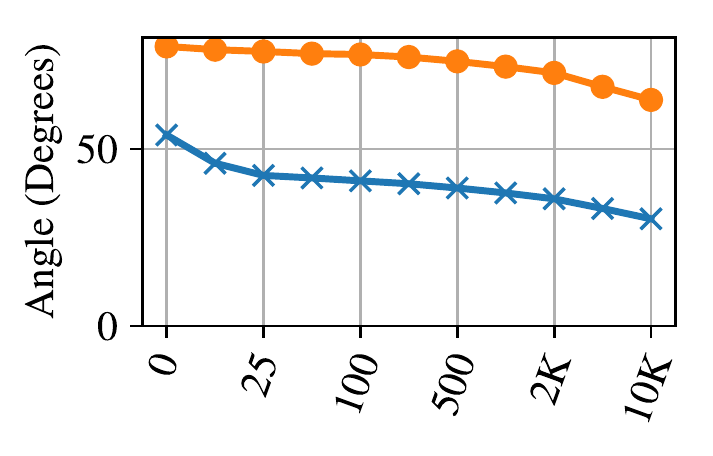}%
\includegraphics[width=0.25\textwidth]{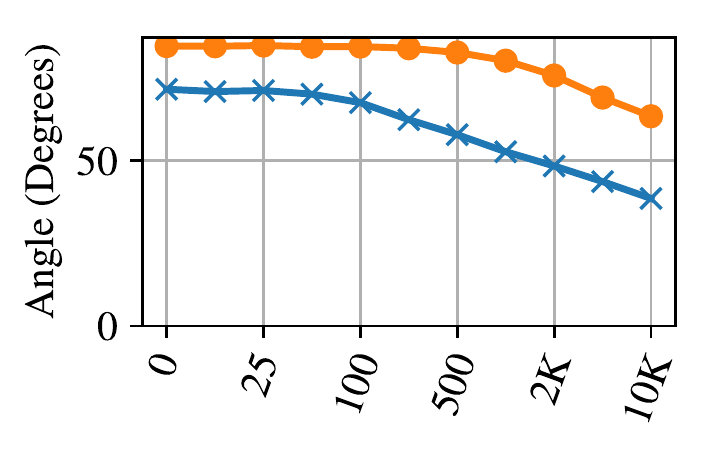}

\vspace{-.6em}%
\includegraphics[width=0.25\textwidth]{figures/late/vgg19-accuracy-dataorder}%
\includegraphics[width=0.25\textwidth]{figures/late/vgg19-low-accuracy-dataorder}%
\includegraphics[width=0.25\textwidth]{figures/late/vgg19-warmup-accuracy-dataorder}

\caption{The effect of the rewinding iteration (x-axis) on angle data order stability (top), angle pruning stability (middle), and accuracy (bottom) for each network in Figure \ref{fig:small-networks}. This figure accompanies Figure \ref{fig:rewinding}.}
\label{fig:rewinding-angle}
\end{figure}

\begin{figure}
\vspace{-.4em}%
\includegraphics[width=0.25\textwidth]{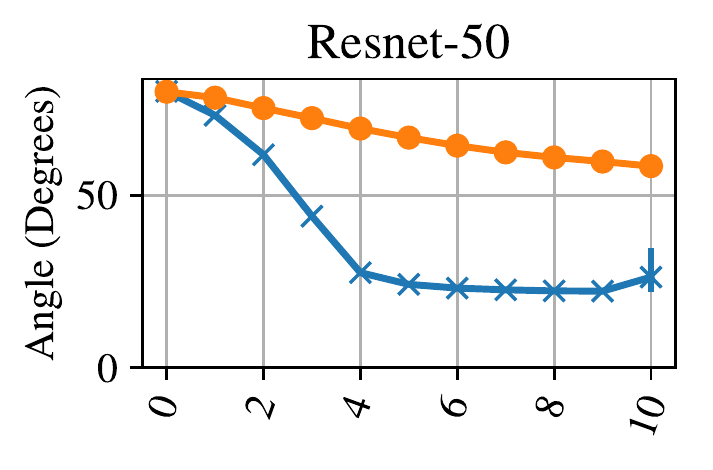}%
\includegraphics[width=0.25\textwidth]{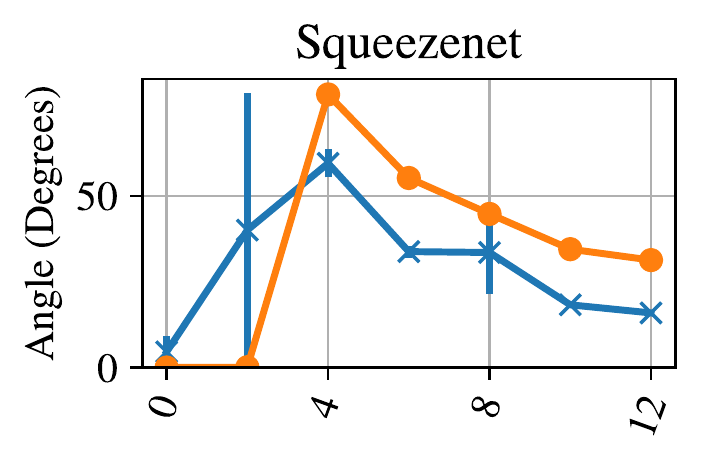}%
\includegraphics[width=0.25\textwidth]{figures/late/inception-parameter-dataorder}
\includegraphics[width=.2\textwidth]{figures/late/lenet-accuracy-dataorder-legend}\hspace{.05\textwidth}%

\vspace{-.6em}
\includegraphics[width=0.25\textwidth]{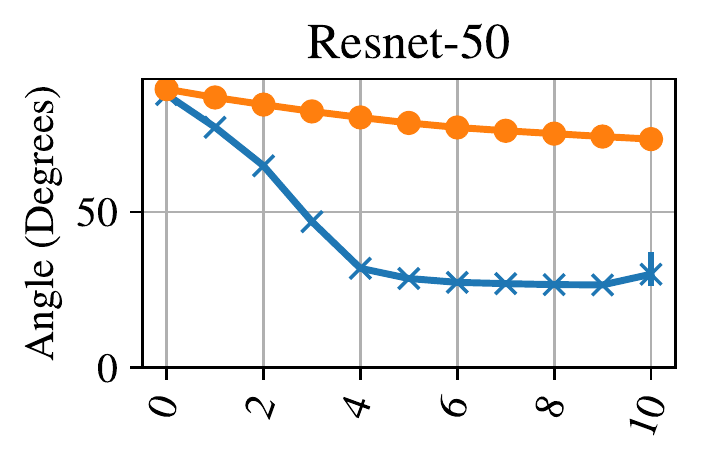}%
\includegraphics[width=0.25\textwidth]{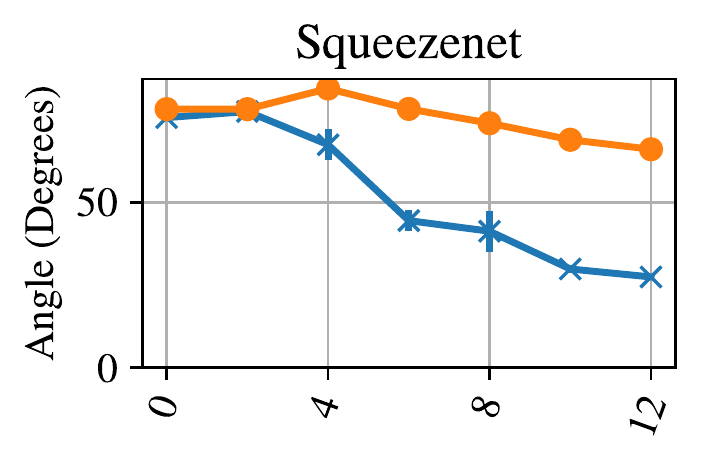}%
\includegraphics[width=0.25\textwidth]{figures/late/inception-parameter-dataorder}

\vspace{-.6em}%
\includegraphics[width=0.25\textwidth]{figures/late/resnet50-accuracy-dataorder}%
\includegraphics[width=0.25\textwidth]{figures/late/squeezenet-accuracy-dataorder}%
\includegraphics[width=0.25\textwidth]{figures/late/inception-accuracy-dataorder}

\caption{The effect of the rewinding epoch (x-axis) on data order stability (top), pruning stability (middle), and accuracy (bottom) for each network in Figure \ref{fig:large-networks}.}
\label{fig:large-angle}
\end{figure}

This appendix accompanies Figures \ref{fig:stability-at-zero} and \ref{fig:rewinding}, which measure the stability of the networks in Figure \ref{fig:small-networks}.
The aforementioned figures measure stability only in terms of distance.
This appendix includes the accompanying measurements of angle.
Figure \ref{fig:stability-at-zero-angle} includes angle data when resetting at iteration 0 to accompany Figure \ref{fig:stability-at-zero}.
Figure \ref{fig:rewinding-angle} includes the angle measurements of stability to data order (top) and pruning (middle) when networks are rewound to various iterations; it accompanies Figure \ref{fig:rewinding}.
Figure \ref{fig:large-angle} includes angle measurements for the ImageNet networks from Section \ref{sec:imagenet} to accompany Figure \ref{fig:imagenet-late-resetting}.

\end{appendix}

\end{document}